%% file: main.tex
\documentclass[10pt,twocolumn,letterpaper]{article}
\pdfoutput=1

\usepackage{cvpr}
\usepackage{times}
\usepackage{epsfig}
\usepackage{graphicx}
\usepackage{amsmath}
\usepackage{amssymb}
\usepackage{caption}
\usepackage{subcaption}

\usepackage{dsfont}

\usepackage[pagebackref=true,breaklinks=true,letterpaper=true,colorlinks,bookmarks=false]{hyperref}

\newcommand{\denselist}{
\itemsep -2pt\topsep-8pt\partopsep-8pt
}

 \cvprfinalcopy 

\newcommand{\eat}[1]{}

\newcommand{\todo}[1]{}
\newcommand{\Jon}[1]{}
\newcommand{\Kevin}[1]{}

\input{macros}

\renewcommand{\paragraph}[1]{\subsubsection{#1}}

\begin{document}

\title{Speed/accuracy trade-offs for modern convolutional object detectors}

\author{Jonathan Huang
  \and
  Vivek Rathod
  \and
  Chen Sun
  \and
  Menglong Zhu
  \and
  Anoop Korattikara
  \and
  Alireza Fathi
  \and
  Ian Fischer
  \and
  Zbigniew  Wojna
  \and
  Yang Song
  \and
  Sergio Guadarrama
\and
Kevin Murphy\\
Google Research
}

\maketitle

\input{abstract}

\input{intro}
\input{related}

\input{methods}

\input{results}

\input{concl}



\subsubsection*{Acknowledgements}\vspace{-3mm}
{\footnotesize
We would like to thank the following people
for their advice and support throughout this project: 
Tom Duerig, Dumitru Erhan, Jitendra Malik, George Papandreou, Dominik Roblek,
Chuck Rosenberg, Nathan Silberman, Abhinav Srivastava, Rahul Sukthankar,
Christian Szegedy, Jasper Uijlings, Jay Yagnik, Xiangxin Zhu.
}

{\small
\bibliographystyle{abbrv}
\bibliography{main}
}

\end{document}


\title{Speed/accuracy trade-offs for modern convolutional object detectors: Supplementary Materials}
\maketitle

This is the supplementary materials document for ``Speed/accuracy trade-offs for modern convolutional object detectors''.
In Section~\ref{sec:model_details}, 
we provide additional details regarding model specification and training. 
In Section~\ref{sec:extra_analyses}, we discuss additional analyses which may be interesting for readers but did not make
it into the main paper.
Separately,  we also include a table containing 
all data collected in this study and a paper that introduces the MobileNet feature extractor under concurrent submission
to CVPR.

\input{model_details.tex}

\input{extra_analyses.tex}

{\small
\bibliographystyle{ieee}
\bibliography{bib}
}

%% file: macros.tex
\usepackage{amsmath}
\usepackage{amssymb}
\usepackage{amsthm}



%% file: abstract.tex
\begin{abstract}\vspace{-3mm}
  The goal of this paper is to serve as a guide for selecting a
   detection architecture that achieves the right
   speed/memory/accuracy balance for a given application and platform.
   To this end, we investigate various ways to trade accuracy for
   speed and memory usage in modern convolutional object detection
   systems.  A number of successful systems have been proposed in
   recent years, but apples-to-apples comparisons are difficult due to
   different base feature extractors (e.g., VGG, Residual Networks),
   different default image resolutions, as well as different hardware
   and software platforms.  We present a unified implementation of the
   Faster R-CNN~\cite{ren2015faster}, R-FCN~\cite{dai2016r} and
   SSD~\cite{liu2015ssd} systems, which we view as
   ``meta-architectures'' and trace out the speed/accuracy trade-off
   curve created by using alternative feature extractors and varying
   other critical parameters such as image size within each of these
   meta-architectures.  On one extreme end of this spectrum where
   speed and memory are critical, we present a detector that achieves
   real time speeds and can be deployed on a mobile device.  On the
   opposite end in which accuracy is critical, we present a detector
   that achieves state-of-the-art performance measured on the COCO
   detection task.
\end{abstract}

%% file: intro.tex
\vspace{-5mm}
\section{Introduction}\vspace{-1mm}
\label{sec:intro}

A lot of progress has been made in recent years on object detection
due to the use of convolutional neural networks (CNNs).  Modern object
detectors based on these networks --- such as
Faster R-CNN~\cite{ren2015faster},
R-FCN~\cite{dai2016r},
Multibox~\cite{szegedy2014scalable},
SSD~\cite{liu2015ssd} and
YOLO~\cite{redmon2015you} ---
are now good enough to be deployed in
consumer products (e.g., Google Photos, Pinterest Visual Search) and
some have been shown to be fast enough to be run on mobile devices.

However,
it can be difficult for practitioners to decide what architecture is best suited
to their application.  Standard accuracy metrics, such as mean average precision
(mAP), do not tell the entire story, since for real
deployments of computer vision systems, running time and memory usage
are also critical. For example,
mobile devices often require a small memory
footprint, and self driving cars require real time performance.
Server-side
production systems, like those used in Google, Facebook or
Snapchat,  have more leeway to optimize for accuracy,
but are still subject to throughput constraints.  While the methods that
win competitions, such as the COCO challenge
\cite{Lin2014coco},
are optimized for accuracy, they often rely on model ensembling and
multicrop methods which are too slow for practical usage.

\eat{
Choosing a detector is also not as simple as pointing to a single
paper (e.g., Faster R-CNN or YOLO) since many of these approaches are
better thought of as a choice of meta-architecture which requires a
number of further choices such as which base feature extractor to use
(e.g., VGG versus Resnet).  From this perspective, there exist a
number of tricks that can be played to trade off speed for accuracy
(e.g. by replacing feature extractors with less resource hungry
feature extractors or training the detectors on smaller input images).
Additionally, in many cases, the practitioner might care more about
accuracy for large objects (which are likely to be close to the
sensor) or for certain classes (such as people and cars for autonomous
driving).
}

Unfortunately, 
only a small subset of papers
(e.g., 
R-FCN~\cite{dai2016r},
SSD~\cite{liu2015ssd}
YOLO~\cite{redmon2015you})
discuss running time in any detail. Furthermore, these papers
typically only state that they achieve some frame-rate, but  do not give
a full picture of the speed/accuracy trade-off, which depends on many
other factors, such as which feature extractor is used, input image sizes,
etc.

In this paper, we seek to explore the speed/accuracy trade-off of
modern detection systems in an exhaustive and fair way.  While this has
been studied for full image classification( (e.g.,~\cite{canziani2016analysis}),
detection models tend to be significantly more complex.
We primarily
investigate single-model/single-pass detectors, by which we mean
models that do not use ensembling, multi-crop methods, or other
``tricks'' such as horizontal flipping.  In other words,
we only pass a single image through a single network.
For simplicity (and because it is more important for users of this
technology), we focus only on test-time performance and not on how
long these models take to train.

Though it is impractical to compare every recently proposed detection
system, we are fortunate that many of the
leading state of the art approaches have converged
on a common methodology
(at least at a high level).
This has allowed us to implement and compare a
large number of detection systems in a unified manner.  In particular,
we have created implementations of the Faster R-CNN, R-FCN and SSD
meta-architectures, which at a high level consist of a single
convolutional network,
trained with a mixed regression and classification objective,
and use sliding window style predictions.

To summarize,
our main contributions are as follows:\vspace{-2mm}
\begin{itemize}
\denselist
\item We provide a concise survey of modern convolutional detection systems, and describe
how the leading ones follow very similar designs.  

\item We describe our flexible and unified implementation of three
meta-architectures (Faster R-CNN, R-FCN and SSD) in Tensorflow which we
use to do extensive experiments that trace the accuracy/speed
trade-off curve for different detection systems, varying
meta-architecture, feature extractor, image resolution, etc.

\item Our findings show that using fewer proposals for Faster R-CNN can speed it up significantly without a big loss in accuracy, making it competitive with its faster cousins, SSD and RFCN.  We show that SSD’s performance is less sensitive to the quality of the feature extractor than Faster R-CNN and R-FCN. And we identify “sweet spots” on the accuracy/speed trade-off curve where gains in accuracy are only possible by sacrificing speed (within the family of detectors presented here).

\item Several of the meta-architecture and feature-extractor
combinations that we report have never appeared before in
literature.  We discuss how we used some of these novel combinations
to 
train the winning entry of the 2016 COCO object detection
challenge. 
\end{itemize}


%% file: related.tex
\section{Meta-architectures}
\label{sec:related}

Neural nets have become the leading method for high quality object detection in recent years. 
In this section we survey some of the highlights of this literature.
The R-CNN paper by Girshick et al.~\cite{girshick2014rich} was among the first modern incarnations of convolutional network based detection.  Inspired by recent successes on image classification~\cite{krizhevsky2012imagenet}, the R-CNN method took the straightforward approach of cropping externally computed box proposals out of an input image and running a neural net classifier on these crops.  This approach can be expensive however because many crops are necessary, leading to significant duplicated computation from overlapping crops.  Fast R-CNN~\cite{girshick2015fast} alleviated this problem by pushing the entire image once through a feature extractor then cropping from an intermediate layer so that crops share the computation load of
feature extraction.

\looseness -1 While both R-CNN and Fast R-CNN relied on an external proposal generator, recent  works have shown that it is possible to generate box proposals using neural networks as well~\cite{szegedy2013deep,szegedy2014scalable,erhan2014scalable,ren2015faster}.  In these works, it is typical to have a collection of boxes overlaid on the image at different spatial locations, scales and aspect ratios that act as ``anchors'' (sometimes called ``priors'' or ``default boxes'').
A model is then trained to make two predictions for each anchor: (1) a discrete class prediction for each anchor, and (2) a continuous prediction of an offset by which the anchor needs to be shifted to fit the groundtruth bounding box.  

\eat{
\begin{figure}[t!]
\begin{center}
\includegraphics[width=0.40\linewidth]{Figures/boat_example.pdf}
\caption{
\footnotesize An example tiled grid of anchors overlaid on an image with
a positive anchor (blue) and its matching groundtruth box (green).  In this image, all other anchors are negative.  Note that in general, anchors overlap and might be tiled in scale and aspect ratio as well. 
}
\end{center}\vspace{-4mm}
\label{fig:boat_example}
\end{figure}
}

\looseness -1 Papers that follow this anchors methodology then minimize a combined classification and regression loss that we now describe.
For each anchor $a$, we first find the best matching groundtruth box $b$ (if one exists).  If such a match can be found, we call $a$ a ``positive anchor'',
and assign it (1) a class label $y_a \in {\{1 \dots K\}}$
and (2) a vector encoding of box $b$ with respect to anchor $a$
(called the box encoding $\phi(b_a; a)$).
If no match is found, we call $a$ a ``negative anchor'' and we set the 
class label to be $y_a=0$.
If for the anchor $a$ we predict box encoding $f_{loc}(\mathcal{I}; a, \theta)$  and corresponding class $f_{cls}(\mathcal{I}; a, \theta)$,
where $\mathcal{I}$ is the image and $\theta$ the model parameters,
then the loss for $a$ is measured as a weighted sum of a location-based loss and a classification loss:\vspace{-3mm}

{\footnotesize
\begin{align}
\mathcal{L}(a, \mathcal{I}; \theta) =&\; \alpha\cdot \mathds{1}[a \,\mbox{is positive}]\cdot \ell_{loc} (\phi(b_a; a) - f_{loc}(\mathcal{I}; a, \theta)) \nonumber \\
 & + \beta\cdot \ell_{cls} (y_a, f_{cls}(\mathcal{I}; a, \theta)),\label{eqn:loss}
\end{align}
}
where $\alpha,\beta$ are weights balancing localization and classification losses.  To train the model, Equation~\ref{eqn:loss} is averaged over anchors and minimized with respect to parameters $\theta$.

The choice of anchors has significant implications both for accuracy and computation.  In the (first) Multibox paper~\cite{erhan2014scalable}, these anchors (called ``box priors'' by the authors) were generated by clustering groundtruth boxes in the dataset.  In more recent works, anchors are generated by tiling a collection of boxes at different scales and aspect ratios regularly across the image.  The advantage of having a regular grid of anchors is that predictions for these boxes can be written as tiled predictors on the image with shared parameters (i.e., convolutions) and are reminiscent of traditional sliding window methods, e.g.~\cite{viola2004robust}.  The Faster R-CNN~\cite{ren2015faster} paper and the (second) Multibox paper~\cite{szegedy2014scalable} (which called these tiled anchors ``convolutional priors'') were the first papers to take this new approach.

\begin{table*}
\begin{center}
{\footnotesize
\begin{tabular}{c|c|c|c|c|c}
Paper & Meta-architecture & Feature Extractor & Matching & Box Encoding $\phi(b_a, a)$ & Location Loss functions \\
\hline
Szegedy et al.~\cite{szegedy2014scalable} & SSD & InceptionV3 & Bipartite & $[x_0, y_0, x_1, y_1]$ & $L_2$ \\
Redmon et al.~\cite{redmon2015you} & SSD & Custom (GoogLeNet inspired) & Box Center  & $[x_c, y_c, \sqrt{w}, \sqrt{h}]$ & $L_2$\\
Ren et al.~\cite{ren2015faster} & Faster R-CNN & VGG  & Argmax & $[\frac{x_c}{w_a}, \frac{y_c}{h_a}, \log w, \log h]$ & Smooth$L_1$\\
He et al.~\cite{he2015deep} & Faster R-CNN & ResNet-101  & Argmax & $[\frac{x_c}{w_a}, \frac{y_c}{h_a}, \log w, \log h]$ & Smooth$L_1$\\
Liu et al.~\cite{liu2015ssd} (v1) & SSD & InceptionV3  & Argmax & $[x_0, y_0, x_1, y_1]$  & $L_2$\\
Liu et al.~\cite{liu2015ssd} (v2, v3) & SSD & VGG  & Argmax & $[\frac{x_c}{w_a}, \frac{y_c}{h_a}, \log w, \log h]$ & Smooth$L_1$ \\
Dai et al~\cite{dai2016r} & R-FCN & ResNet-101  & Argmax & $[\frac{x_c}{w_a}, \frac{y_c}{h_a}, \log w, \log h]$ & Smooth$L_1$ \\
\end{tabular}
}
\vspace{1mm}
\caption{
\footnotesize Convolutional  detection models that use one of the meta-architectures described in Section~\ref{sec:related}.  Boxes are encoded with respect to a matching anchor $a$ via a
function $\phi$ (Equation~\ref{eqn:loss}), where $[x_0, y_0, x_1, y_1]$ are min/max coordinates of a box, $x_c, y_c$ are its center coordinates, and $w, h$ its width and height.  In some cases, $w_a, h_a$,  width and height of the matching anchor are also used.
{\bf Notes}: (1) We include an early arXiv version of \cite{liu2015ssd}, which used a different configuration from that published at ECCV 2016; (2) \cite{redmon2015you} uses a fast feature extractor described as being inspired by GoogLeNet~\cite{szegedy2015going}, which we do not compare to; (3) YOLO matches a groundtruth box to an anchor if its center falls inside the anchor (we refer to this as \emph{BoxCenter}).
}\vspace{-6mm}
\label{tab:paper_comparison}
\end{center}
\end{table*}

\begin{figure*}[t!]
\begin{center}
\begin{subfigure}[t]{0.3\linewidth}
  \includegraphics[width=\linewidth]{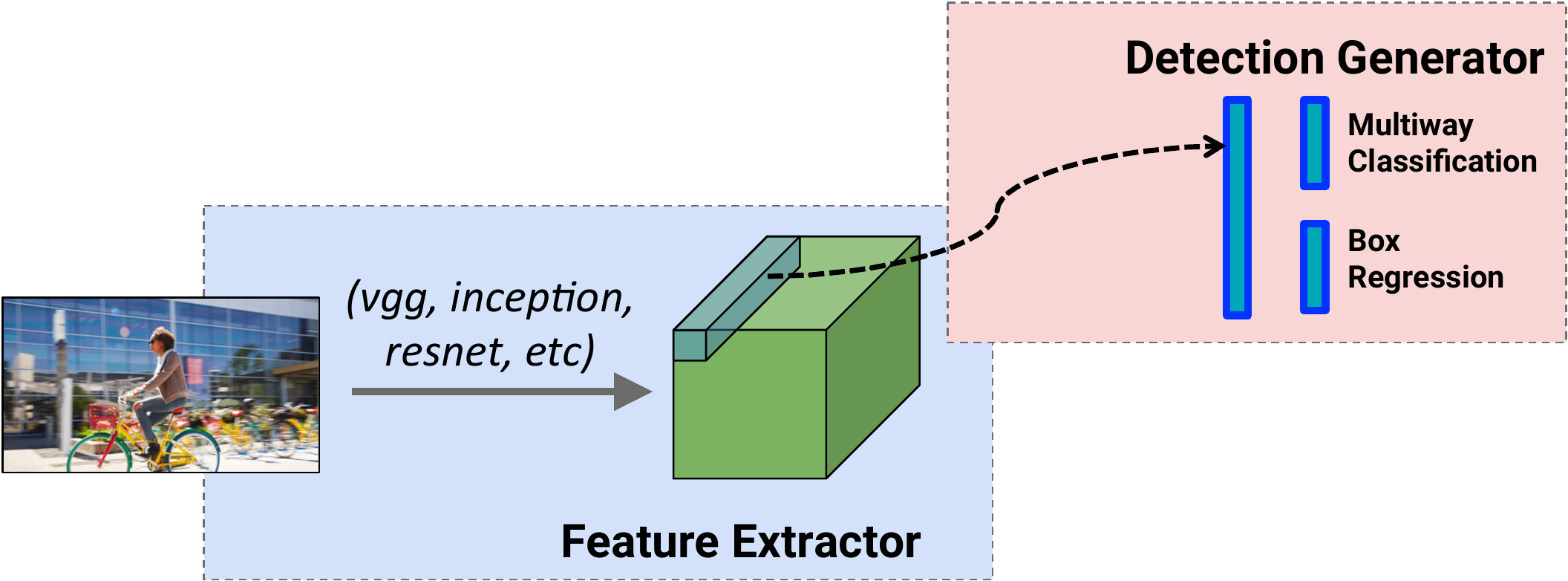}
  \caption{SSD.}
\label{fig:ssd_diagram}
\end{subfigure}
\begin{subfigure}[t]{0.3\linewidth}
  \includegraphics[width=\linewidth]{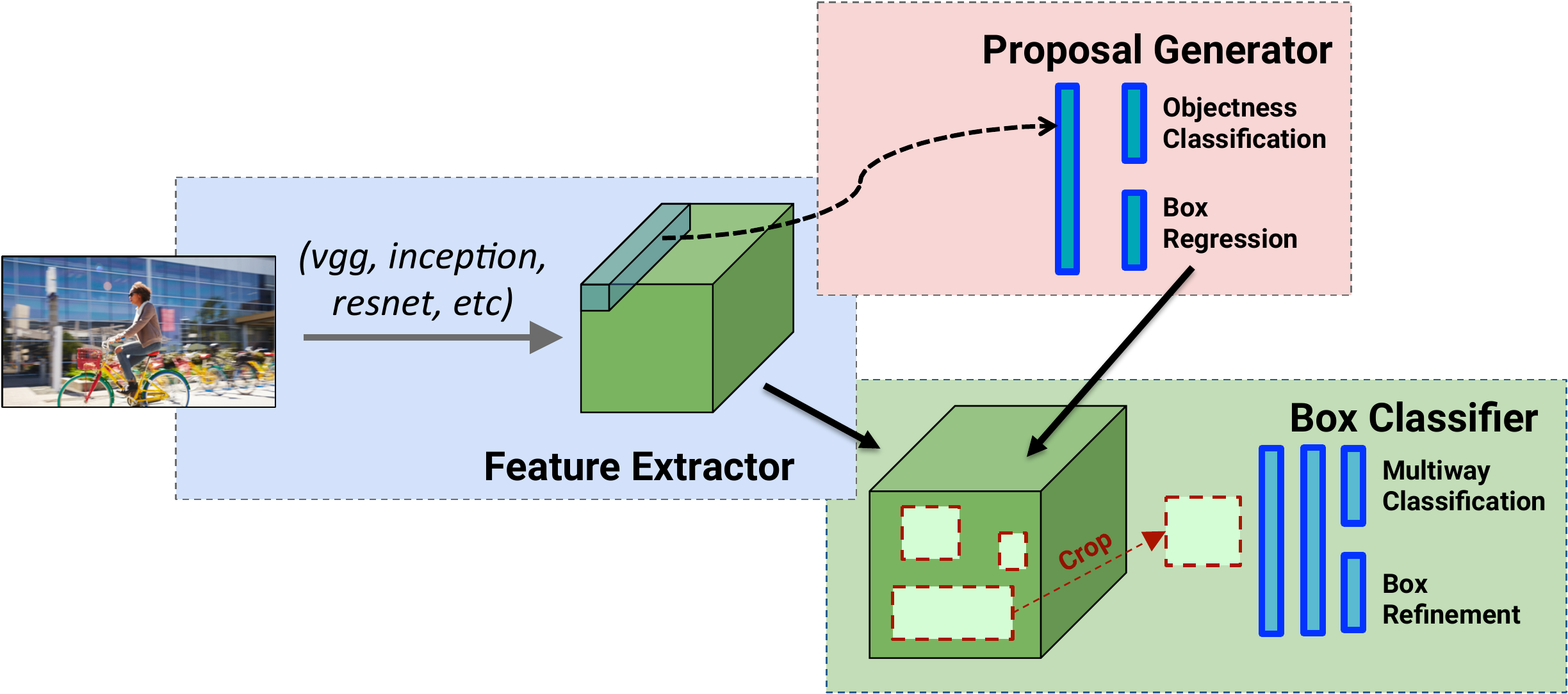}
  \caption{Faster RCNN.}
\label{fig:fasterrcnn_diagram}
\end{subfigure}
\begin{subfigure}[t]{0.3\linewidth}
  \includegraphics[width=\linewidth]{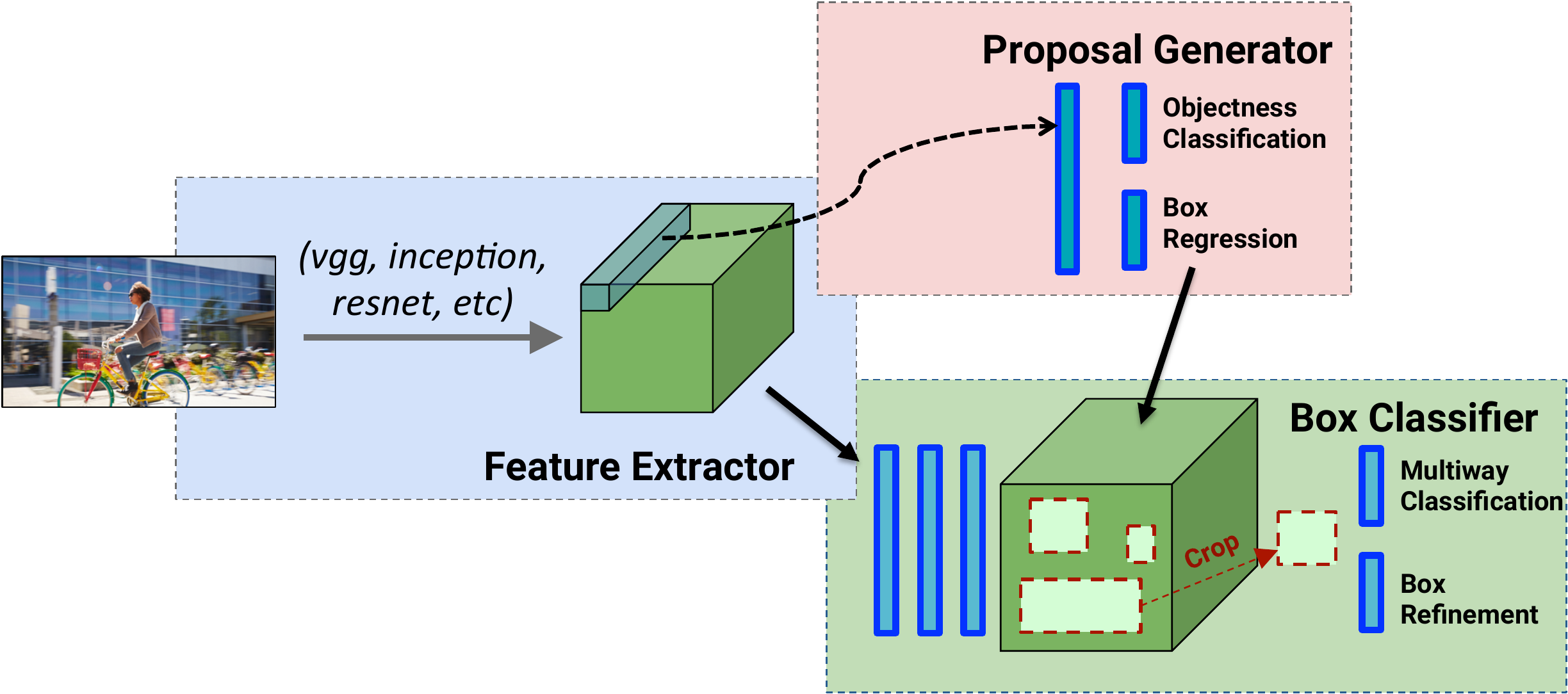}
  \caption{R-FCN.}
\label{fig:rfcn_diagram}
\end{subfigure}
\caption{
High level diagrams of the detection meta-architectures compared in this paper.
}
\end{center}
\end{figure*}

\subsection{Meta-architectures}\vspace{-2mm}
In our paper we focus primarily on three recent  (meta)-architectures: SSD (Single Shot Multibox Detector~\cite{liu2015ssd}), Faster R-CNN~\cite{ren2015faster} and R-FCN (Region-based Fully Convolutional Networks~\cite{dai2016r}).
While these papers were originally presented with a particular feature extractor (e.g., VGG, Resnet, etc), we now review these three methods, decoupling the choice of meta-architecture from feature extractor so that conceptually, any feature extractor can be used with SSD, Faster R-CNN or R-FCN.
\vspace{-3mm}
\paragraph{Single Shot Detector (SSD).}
Though the \emph{SSD} paper was published only recently (Liu et al.,~\cite{liu2015ssd}),
we use the term SSD to refer broadly to architectures that use a single feed-forward
convolutional network to directly predict classes
and anchor offsets without requiring a second stage per-proposal classification operation
(Figure~\ref{fig:ssd_diagram}).  Under this definition, the SSD meta-architecture has been explored in a number of precursors to~\cite{liu2015ssd}. Both Multibox and the Region Proposal Network (RPN) stage of Faster R-CNN~\cite{szegedy2014scalable,ren2015faster} use this approach to predict class-agnostic box proposals.  \cite{sermanet2013overfeat,redmon2015you,redmon2016yolo9000,fu2017dssd} use SSD-like architectures to predict final (1 of $K$) class labels.  And Poirson et al.,~\cite{poirson2016fast} extended this idea to predict boxes, classes and pose.


\vspace{-3mm}
\paragraph{Faster R-CNN.}
In the Faster R-CNN setting, detection happens in two stages 
(Figure~\ref{fig:fasterrcnn_diagram}). 
In the first stage, called the \emph{region proposal network} (RPN), images are processed by a feature extractor (e.g., VGG-16), and features at some selected intermediate level (e.g., ``conv5'') are used to predict class-agnostic box proposals.  The loss function for this first stage takes the form of Equation~\ref{eqn:loss} using a grid of anchors tiled in space, scale and aspect ratio.  

In the second stage, these (typically 300) box proposals are used to crop  features from the same intermediate feature map which are subsequently fed to the remainder of the feature extractor (e.g., ``fc6'' followed by ``fc7'') in order to predict a class and class-specific box refinement for each proposal.  The loss function for this second stage \emph{box classifier} also takes the form of Equation~\ref{eqn:loss} using the proposals generated from the RPN as anchors.  Notably, one does \emph{not} crop proposals directly from the image and re-run crops through the feature extractor, which would be duplicated computation.
However there is part of the computation that must be run once per region, and thus the running time depends on the number of regions proposed by the RPN.

Since appearing in 2015, Faster R-CNN has been particularly influential, and has led to a number of follow-up works~\cite{bell2015inside,shrivastava2016training,shrivastava2016contextual,zagoruyko2016multipath,he2015deep,dai2015instance,kim2016pvanet,yang2016craft,lin2016feature,zhai2017visual} (including SSD and R-FCN).  Notably, half of the submissions to the COCO object detection server as of November 2016 are reported to be based on the Faster R-CNN system in some way.


\subsection{R-FCN}
While Faster R-CNN is an order of magnitude faster than Fast R-CNN, the fact that the region-specific component must be applied several hundred times per image led Dai et al.~\cite{dai2016r} to propose the \emph{R-FCN} (Region-based Fully Convolutional Networks) method which is like Faster R-CNN, but instead of cropping features from the same layer where region proposals are predicted, crops are taken from the last layer of features prior to prediction (Figure~\ref{fig:rfcn_diagram}).  This approach of pushing cropping to the last layer minimizes the amount of per-region computation that must be done.
Dai et al. argue that the object detection task needs localization representations that respect translation variance and thus propose a position-sensitive cropping mechanism that is used instead of the more standard ROI pooling operations used in ~\cite{girshick2015fast,ren2015faster} and the differentiable crop mechanism of ~\cite{dai2015instance}.  They show that the R-FCN model (using Resnet 101) could achieve comparable accuracy to Faster R-CNN often at faster running times.  Recently, the R-FCN model was also adapted to do instance segmentation in the recent \emph{TA-FCN model}~\cite{li16}, which won the 2016 COCO instance segmentation challenge.


\Jon{we are missing the Gidaris papers}

%% file: methods.tex
\section{Experimental setup}
\label{sec:methods}
\vspace{-2mm}
The introduction of standard benchmarks such as Imagenet~\cite{russakovsky2015imagenet} and COCO~\cite{Lin2014coco} has made it easier in recent years to compare detection methods with respect to accuracy.  However, when it comes to speed and memory, apples-to-apples comparisons have been harder to come by. Prior works have relied on different deep learning frameworks (e.g., DistBelief~\cite{dean2012large}, Caffe~\cite{jia2014caffe}, Torch~\cite{collobert2011torch7}) and different hardware.  Some papers have optimized for accuracy; others for speed.  
And finally, in some cases, metrics are reported using slightly different training sets (e.g., COCO training set vs. combined training+validation sets).

In order to better perform apples-to-apples comparisons, we have created a   detection platform in Tensorflow~\cite{abadi2015tensorflow} and have recreated training pipelines for SSD, Faster R-CNN and R-FCN meta-architectures on this platform.  Having a unified framework has allowed us to easily swap feature extractor architectures, loss functions, 
and having it in Tensorflow allows for easy portability to diverse platforms for deployment.  In the following we discuss ways to configure model architecture, loss function and input on our platform --- knobs that can be used to trade speed and accuracy.

\subsection{Architectural configuration}

\paragraph{Feature extractors.}
In all of the meta-architectures, we first apply a convolutional \emph{feature extractor} to the input image to obtain high-level features.  The choice of feature extractor is crucial as the number of parameters and types of layers directly affect memory, speed, and performance of the detector.
We have selected six representative feature extractors to compare in this paper
and, with the exception of MobileNet \cite{howard17}, all have open source  Tensorflow implementations and have had sizeable influence on the vision community. 

In more detail, we consider the following six feature extractors.
We use
{\bf VGG-16}~\cite{simonyan2014very}
and {\bf  Resnet-101}~\cite{he2015deep}, both of
which have won many competitions such as ILSVRC and COCO 2015
(classification, detection and segmentation).
We also use  {\bf Inception v2}~\cite{ioffe2015batch},
which set the state of the art in the ILSVRC 2014 classification and
detection challenges,
as well as its successor {\bf Inception  v3}~\cite{szegedy2015rethinking}.
Both of the Inception networks
employed `Inception units' which made it possible to increase the
depth and width of a network without increasing its computational
budget.  Recently, Szegedy et al.~\cite{szegedy2016inception} proposed
{\bf Inception Resnet (v2)}, which combines the optimization benefits
conferred by residual connections with the computation efficiency of
Inception units.  Finally, we compare against the new {\bf MobileNet}
network~\cite{howard17}, which has been shown to achieve VGG-16 level
accuracy on Imagenet with only $1/30$ of the computational cost
and model size. MobileNet is designed for efficient inference in
various mobile vision applications. Its building blocks are depthwise
separable convolutions which factorize a standard convolution into a
depthwise convolution and a $1 \times 1$ convolution,  
effectively reducing both computational cost and number of
parameters.



For each feature extractor, there are choices to be made in order to use it within a meta-architecture.  For both Faster R-CNN and R-FCN, one must choose which layer to use for predicting region proposals.  In our experiments, we use the choices laid out in the original papers when possible.  For example, we use the `conv5' layer from VGG-16~\cite{ren2015faster} and the last layer of conv\_4\_x layers in Resnet-101~\cite{he2015deep}.  For other feature extractors, we have made analogous choices.  See supplementary materials for more details.

Liu et al.~\cite{liu2015ssd} showed that in the SSD setting,
using multiple feature maps to make location and confidence 
predictions at multiple scales is critical for good performance.
For VGG feature extractors, they used conv4\_3, fc7 (converted to a convolution layer),
as well as a sequence of added layers.  In our experiments, we follow
their methodology closely, always selecting the topmost convolutional feature map and a 
higher resolution feature map at a lower level, then adding a sequence of 
convolutional layers with spatial resolution decaying by a factor of 2
with each additional layer used for prediction.  However unlike~\cite{liu2015ssd}, we use batch normalization in all additional layers.

For comparison, feature extractors used in previous works are shown in Table \ref{tab:paper_comparison}. In this work, we evaluate all combinations of meta-architectures and feature extractors, most of which are novel. Notably, Inception networks have never been used in Faster R-CNN frameworks and until recently were not open sourced~\cite{silberman16}. Inception Resnet (v2) and MobileNet have not appeared in the detection literature to date. 

\paragraph{Number of proposals.}
For Faster R-CNN and R-FCN, we can also choose the number of region proposals to be sent to the box classifier at test time. 
Typically, this number is 300 in both settings, but an easy way to save computation is to send fewer boxes
potentially at the risk of reducing recall.  In our experiments, we vary this number of proposals between 10 and 300 in order
to explore this trade-off.

\paragraph{Output stride settings for Resnet and Inception Resnet.}
Our implementation of Resnet-101 is slightly modified from the
original to have an
effective output stride of 16 instead of 32;
we achieve this by modifying the conv5\_1
layer to have stride 1 instead of  2 (and compensating for reduced
stride by using atrous convolutions in further layers) as
in~\cite{dai2016r}.  For Faster R-CNN and R-FCN, in addition to the
default stride of 16, we also experiment with a (more expensive)
stride 8 Resnet-101 in which the conv4\_1 block is additionally
modified to have stride 1.  Likewise, we experiment with stride 16 and
stride 8 versions of the Inception Resnet network.
We find that using stride 8 instead of 16
improves the mAP by a factor of 5\%\footnote{
  i.e., (map8 - map16) / map16 = 0.05.
  },
but increased running time by a factor of 63\%.


\subsection{Loss function configuration}
Beyond selecting a feature extractor, there are choices in configuring the loss function (Equation~\ref{eqn:loss}) which can impact training stability and final performance. Here we describe the choices that we have made in our experiments and Table~\ref{tab:paper_comparison} again compares how similar loss functions are configured in other works.

\paragraph{Matching.} 
Determining classification and regression targets for 
each anchor requires matching anchors to groundtruth instances.  Common approaches include greedy bipartite matching (e.g., based on Jaccard overlap) or  many-to-one matching strategies in which bipartite-ness is not required, but matchings are discarded if Jaccard overlap between an anchor and groundtruth is too low.  We refer to these strategies as \emph{Bipartite}
or \emph{Argmax}, respectively.  In our experiments we use \emph{Argmax} matching throughout with thresholds set as suggested in the original paper for each meta-architecture. After matching, there is typically a sampling procedure designed to bring the number of positive anchors and negative anchors to some desired ratio.  In our experiments, we also fix these ratios to be those recommended by the paper for each meta-architecture.

\paragraph{Box encoding.} To encode a groundtruth box with respect to its matching anchor, we use the box encoding function $\phi(b_a; a)=[10\cdot\frac{x_c}{w_a}, 10\cdot\frac{y_c}{h_a}, 5\cdot\log w, 5\cdot\log h]$ (also used by~\cite{girshick2014rich,girshick2015fast,ren2015faster,liu2015ssd}). Note that the scalar multipliers 10 and 5 are typically used in all of these prior works, even if not explicitly mentioned.

\paragraph{Location loss ($\ell_{loc}$).} Following~\cite{girshick2015fast,ren2015faster,liu2015ssd}, we use the Smooth L1 (or Huber~\cite{huber1964robust}) loss function in all experiments.

\subsection{Input size configuration.}
In Faster R-CNN and R-FCN, models are trained on images scaled to $M$
pixels on the shorter edge whereas in SSD,  images are always resized
to a fixed shape $M\times M$.  We explore evaluating each model on
downscaled images as a way to trade accuracy for speed.  In particular, we have trained high and low-resolution versions of each model.  
In the ``high-resolution'' settings, we set $M=600$, and in the ``low-resolution'' setting, we set $M=300$.  In both cases, this means that the SSD method processes  fewer pixels on average than a Faster R-CNN or R-FCN model with all other variables held constant.

\subsection{Training and hyperparameter tuning}
We jointly train all models end-to-end using asynchronous gradient
updates on a distributed cluster~\cite{dean2012large}.  For Faster RCNN and R-FCN, 
we use SGD with momentum with batch sizes of 1 (due to these models being 
trained using different image sizes) and for 
SSD, we use RMSProp~\cite{tieleman2012lecture} with batch sizes of 32 (in a few 
exceptions we reduced the batch size for memory reasons).  Finally we 
manually tune learning rate schedules
individually for each feature extractor.  For the model configurations that match works in literature (\cite{ren2015faster,dai2016r,he2015deep,liu2015ssd}), we have reproduced or surpassed the reported mAP results.\footnote{In the case of SSD with VGG, we have reproduced the number reported in the ECCV version of the paper, but the most recent version on ArXiv uses an improved data augmentation scheme to obtain somewhat higher numbers, which we have not yet experimented with.}

Note that for Faster R-CNN and R-FCN, this end-to-end approach
is slightly different from the 4-stage 
training procedure that is typically used.  Additionally, instead of 
using the ROI Pooling layer and Position-sensitive ROI Pooling layers 
used by~\cite{ren2015faster,dai2016r}, we use Tensorflow's ``crop\_and\_resize''
operation which uses bilinear interpolation to resample part of an image
onto a fixed sized grid.  This is similar to the differentiable cropping
mechanism of~\cite{dai2015instance}, the attention model of~\cite{gregor2015draw}
as well as the Spatial Transformer Network~\cite{jaderberg2015spatial}.  However
we disable backpropagation with respect to bounding box coordinates as we have
found this to be unstable during training.

Our networks are trained on the COCO dataset, using all training images
as well as a subset of validation images, holding out 8000 examples for
validation.\footnote{We remark that this dataset is similar but slightly smaller than
the trainval35k set that has been used in several 
papers, e.g.,~\cite{bell2015inside,liu2015ssd}.}
Finally at test time, we post-process detections with non-max suppression using an IOU
threshold of 0.6 and clip all boxes to the image window.  To evaluate our final detections,
we use the official COCO API~\cite{dollar14}, which measures mAP averaged over IOU
thresholds in [0.5 : 0.05 : 0.95], amongst other metrics.

\eat{
\begin{table}
  \begin{tabular}{ll}
    Meta &     Faster RCNN, R-FCN, SSD \\
    Features & VGG, Resnet-101, Inception v2, Inception-ResNet v2,
    MobileNet \\
    \# Proposals & 10, 30, 50, 100, 300 \\
    Resolution & 300, 600
  \end{tabular}
  \caption{Configurations}
\end{table}
}


\subsection{Benchmarking procedure}
To time our models, we use a machine with 32GB RAM, Intel Xeon E5-1650 v2 processor and an Nvidia GeForce GTX Titan X GPU card. Timings are reported on GPU for a batch size of one. The images used for timing are resized so that the smallest size is at least $k$ and then cropped to $k \times k$ where $k$ is either 300 or 600 based on the model. We average the timings over 500 images.

We include postprocessing in our timing (which includes non-max suppression and currently runs only on the CPU).  Postprocessing can take up the bulk of the running time for the fastest models at $\sim 40$ms and currently caps our maximum framerate to 25 frames per second.
Among other things, this means that while our timing results are comparable amongst each other, they may not be directly comparable to other reported speeds in the literature.
Other potential differences include hardware, software drivers, framework (Tensorflow in our case), and batch size (e.g., the Liu et al.~\cite{liu2015ssd} report timings using batch sizes of 8). 
Finally, we use tfprof~\cite{pan2016tfprof} to measure the total memory demand of the models during inference; this gives a more platform independent measure of memory demand. We also average the memory measurements over three images.

\input{model_details}

%% file: model_details.tex
\subsection{Model Details}\label{sec:model_details}

Table~\ref{tab:features} summarizes the feature extractors that we use.
All models are pretrained on ImageNet-CLS.
We give details on how we train the object detectors using these feature extractors below.

\subsubsection{Faster R-CNN}
We follow the original implementation of Faster RCNN~\cite{ren2015faster} closely, but but use Tensorflow's ``crop\_and\_resize'' operation instead of standard ROI pooling .
Except for VGG, all the feature extractors use batch normalization after convolutional layers. We freeze the batch normalization parameters to be those estimated during ImageNet pretraining.
We train faster RCNN with asynchronous SGD with momentum of 0.9. The initial learning rates depend on which feature extractor we used, as explained below. We reduce the learning rate by 10x after 900K iterations and another 10x after 1.2M iterations. 9 GPU workers are used during asynchronous training. Each GPU worker takes a single image per iteration; the minibatch size for RPN training is 256, while the minibatch size for box classifier training is 64.
\begin{itemize}\denselist
\item {\bf VGG~\cite{simonyan2014very}}: We extract features from the ``conv5'' layer whose stride size is 16 pixels. Similar to~\cite{dai2015instance}, we crop and resize feature maps to 14x14 then maxpool to 7x7. The initial learning rate is 5e-4.
\item {\bf Resnet 101~\cite{he2015deep}}: We extract features from the last layer of the ``conv4'' block. When operating in atrous mode, the stride size is 8 pixels, otherwise it is 16 pixels. Feature maps are cropped and resized to 14x14 then maxpooled to 7x7. The initial learning rate is 3e-4.
\item {\bf Inception V2~\cite{ioffe2015batch}}: We extract features from the ``Mixed\_4e'' layer whose stride size is 16 pixels. Feature maps are cropped and resized to 14x14. The initial learning rate is 2e-4.
\item {\bf Inception V3~\cite{szegedy2015rethinking}}: We extract features from the  ``Mixed\_6e'' layer whose stride size is 16 pixels. Feature maps are cropped and resized to 17x17. The initial learning rate is 3e-4.
\item {\bf Inception Resnet~\cite{szegedy2016inception}}: We extract features the  from ``Mixed\_6a'' layer including its associated residual layers. When operating in atrous mode, the stride size is 8 pixels, otherwise is 16 pixels. Feature maps are cropped and resized to 17x17. The initial learning rate is 1e-3.
\item {\bf MobileNet~\cite{howard17}}: We extract features from the  ``Conv2d\_11'' layer whose stride size is 16 pixels. Feature maps are cropped and resized to 14x14. The initial learning rate is 3e-3.
\end{itemize}

\begin{table}
  \begin{tabular}{lll}
    Model & Top-1 accuracy & Num. Params. \\ \hline
    VGG-16 & 71.0 & 14,714,688 \\
    MobileNet & 71.1 & 3,191,072\\
    Inception V2 & 73.9 & 10,173,112\\
    ResNet-101 & 76.4 & 42,605,504\\
    Inception V3 & 78.0 & 21,802,784 \\
    Inception Resnet V2 & 80.4 & 54,336,736
  \end{tabular}
  \caption{Properties of the 6 feature extractors that we use.
    Top-1 accuracy is the classification accuracy on ImageNet.}
  \label{tab:features}
  \end{table}

\subsubsection{R-FCN}
We follow the implementation of R-FCN~\cite{dai2016r} closely, but use Tensorflow's ``crop\_and\_resize'' operation instead of ROI pooling to crop regions from the position-sensitive score maps. All feature extractors use batch normalization after convolutional layers. We freeze the batch normalization parameters to be those estimated during ImageNet pretraining. 
We train R-FCN with asynchronous SGD with momentum of 0.9. 9 GPU workers are used during asynchronous training. Each GPU worker takes a single image per iteration; the minibatch size for RPN training is 256.
As of the time of this submission, we do not have R-FCN results for VGG or Inception V3
feature extractors.
\begin{itemize}\denselist
\item {\bf Resnet 101~\cite{he2015deep}}: We extract features from ``block3'' layer. When operating in atrous mode, the stride size is 8 pixels, otherwise it is 16 pixels. Position-sensitive score maps are cropped with spatial bins of size 7x7 and resized to 21x21. We use online hard example mining to sample a minibatch of size 128 for training the box classifier. The initial learning rate is 3e-4. It is reduced by 10x after 1M steps and another 10x after 1.2M steps.
\item {\bf Inception V2~\cite{ioffe2015batch}}: We extract features from ``Mixed\_4e'' layer whose stride size is 16 pixels. Position-sensitive score maps are cropped with spatial bins of size 3x3 and resized to 12x12. We use online hard example mining to sample a minibatch of size 128 for training the box classifier. The initial learning rate is 2e-4. It is reduced by 10x after 1.8M steps and another 10x after 2M steps.
\item {\bf Inception Resnet~\cite{szegedy2016inception}}: We extract features from ``Mixed\_6a'' layer including its associated residual layers. When operating in atrous mode, the stride size is 8 pixels, otherwise it is 16 pixels. Position-sensitive score maps are cropped with spatial bins of size 7x7 and resized to 21x21. We use all proposals from RPN for box classifier training. The initial learning rate is 7e-4. It is reduced by 10x after 1M steps and another 10x after 1.2M steps.
\item {\bf MobileNet~\cite{howard17}}: We extract features from ``Conv2d\_11'' layer whose stride size is 16 pixels. Position-sensitive score maps are cropped with spatial bins of size 3x3 and resized to 12x12.  We use online hard example mining to sample a minibatch of size 128 for training the box classifier. The initial learning rate is 2e-3. Learning rate is reduced by 10x after 1.6M steps and another 10x after 1.8M steps.
\end{itemize}

\subsubsection{SSD}
As described in the main paper, we follow
the methodology of \cite{liu2015ssd} closely, 
generating anchors in the same way and
selecting the topmost convolutional feature map and a 
higher resolution feature map at a lower level, then adding a sequence of 
convolutional layers with spatial resolution decaying by a factor of 2
with each additional layer used for prediction. The feature map selection for Resnet101 is slightly different, as described below.

Unlike~\cite{liu2015ssd}, we use batch normalization in all additional layers, and 
initialize weights with a truncated normal distribution with a standard deviation of $\sigma=.03$.
With the exception of VGG, we also do not perform ``layer normalization'' (as suggested in~\cite{liu2015ssd}) as we found it not to be necessary for the other feature extractors.
Finally, we employ distributed training with asynchronous SGD using 11 worker machines.
Below we discuss the specifics for each feature extractor that we have considered.
As of the time of this submission, we do not have SSD results for the Inception V3
feature extractor and we only have results for high resolution SSD models 
using the Resnet 101 and Inception V2 feature extractors.
\begin{itemize}\denselist
\item {\bf VGG~\cite{simonyan2014very}}:
Following the paper, we use conv4\_3, and fc7 layers, appending
five additional convolutional layers with decaying spatial resolution with depths
512, 256, 256, 256, 256, respectively.
We apply $L_2$ normalization to the conv4\_3 layer, scaling the feature norm at each location in the feature map to a learnable scale, $s$, which is initialized to 20.0.

During training, we use a base learning rate of $lr_{base}=.0003$, but 
use a warm-up learning rate scheme in which we first train with a
learning rate of $0.8^2\cdot lr_{base}$ for 10K iterations followed by $0.8\cdot lr_{base}$
for another 10K iterations.

\item {\bf Resnet 101~\cite{he2015deep}}:
We use the feature map from the last layer of the ``conv4'' block. When operating in atrous mode, the stride size is 8 pixels, otherwise it is 16 pixels. Five additional convolutional layers with decaying spatial resolution are appended, which have depths 512, 512, 256, 256, 128, respectively. We have experimented with including the feature map from the last layer of the ``conv5'' block. With ``conv5'' features, the mAP numbers are very similar, but the computational costs are higher. Therefore we choose to use the  last layer of the ``conv4'' block. During training, a base learning rate of 3e-4 is used. We use a learning rate warm up strategy similar to the VGG one. 
  
\item {\bf Inception V2~\cite{ioffe2015batch}}:
We use Mixed\_4c and Mixed\_5c, appending four additional convolutional layers with decaying resolution with depths
512, 256, 256, 128 respectively. We use ReLU6 as the non-linear activation function for each conv layer.
During training, we use a base learning rate of 0.002, followed by learning rate decay of 0.95 every 800k steps.
\item {\bf Inception Resnet~\cite{szegedy2016inception}}:
We use Mixed\_6a and Conv2d\_7b, appending three additional convolutional layers with decaying resolution with depths
512, 256, 128 respectively. We use ReLU as the non-linear activation function for each conv layer.
During training, we use a base learning rate of 0.0005, followed by learning rate decay of 0.95 every 800k steps.
\item {\bf MobileNet~\cite{howard17}}:
We use conv\_11 and conv\_13, appending four additional convolutional layers with decaying resolution with depths
512, 256, 256, 128 respectively. The non-linear activation function we use is ReLU6 and both batch norm parameters
$\beta$ and $\gamma$ are trained. During training, we use a base learning rate of 0.004, followed by learning rate decay
of 0.95 every 800k steps.
\end{itemize}

%% file: results.tex
\section{Results}
\label{sec:results}

\begin{figure*}[t!]
\begin{center}
\includegraphics[width=.9\linewidth]{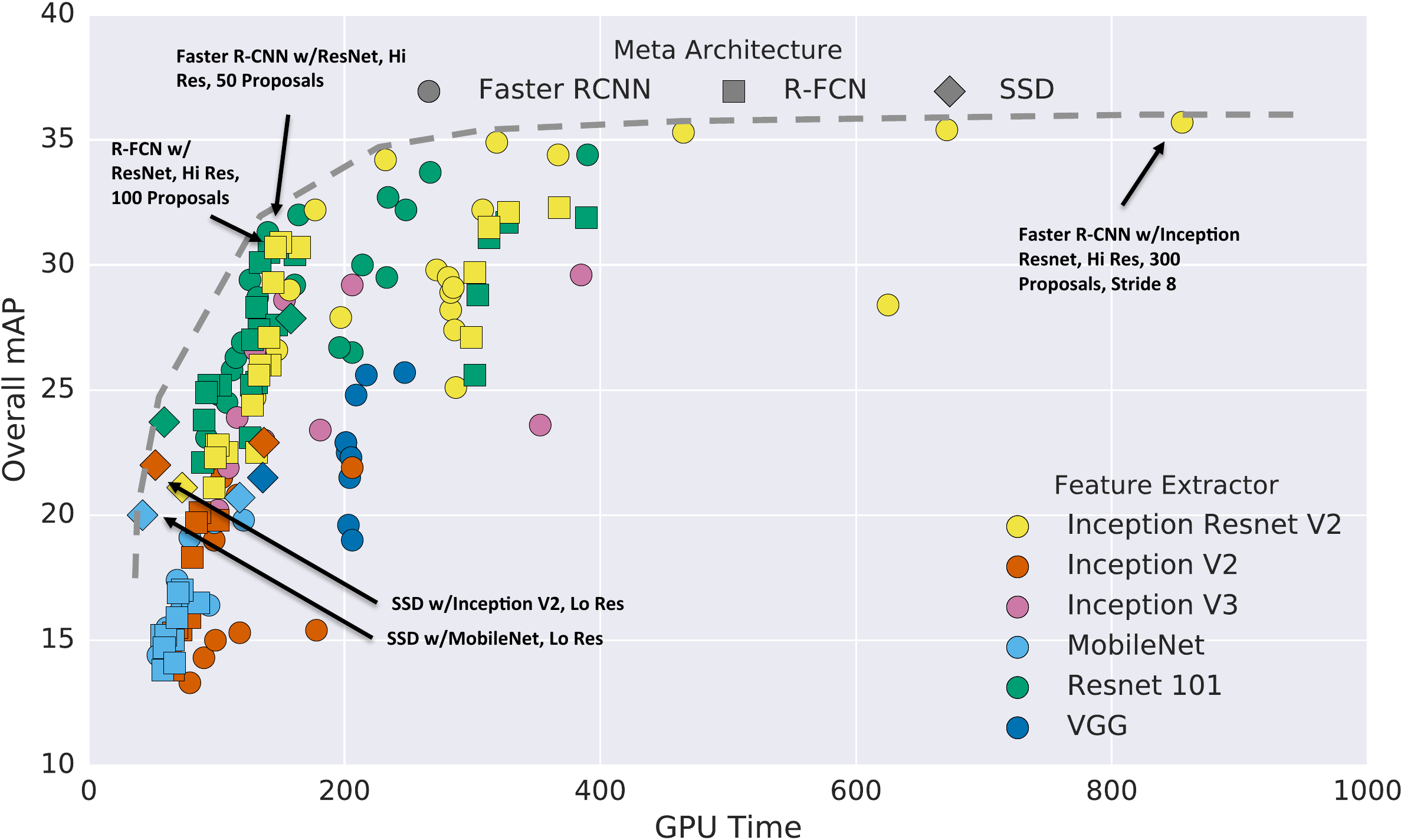}
\caption{
  \footnotesize
Accuracy vs time,
  with marker shapes indicating 
   meta-architecture
  and colors indicating  feature extractor. 
Each (meta-architecture, feature extractor) pair can correspond to multiple points on this plot due to changing input sizes, stride, etc.
}
\label{fig:map_v_gputime_by_meta_and_feat}
\end{center}
\end{figure*}

\begin{table*}[t!]
\begin{center}\footnotesize
\begin{tabular}{c|c|c}
\bf Model summary & \bf minival mAP & \bf test-dev mAP \\
\hline
(Fastest) SSD w/MobileNet (Low Resolution) & 19.3 & 18.8 \\
(Fastest) SSD w/Inception V2 (Low Resolution) & 22 & 21.6 \\
(Sweet Spot) Faster R-CNN w/Resnet 101, 100 Proposals & 32 & 31.9 \\
(Sweet Spot) R-FCN w/Resnet 101, 300 Proposals & 30.4 & 30.3 \\
(Most Accurate) Faster R-CNN w/Inception Resnet V2, 300 Proposals & 35.7 & 35.6\vspace{-3mm}
\end{tabular}
\caption{\footnotesize Test-dev performance of the ``critical'' points along our optimality frontier.}\vspace{-4mm}
\label{tab:testdev}
\end{center}
\end{table*}

In this section we analyze the data that we have collected by training and benchmarking detectors, sweeping over model configurations as described in Section~\ref{sec:methods}.  Each such model configuration includes a choice of meta-architecture, feature extractor, stride (for Resnet and Inception Resnet) as well as input resolution and number of proposals (for Faster R-CNN and R-FCN).

For each such model configuration, we measure timings on GPU, memory demand, number of parameters and floating point operations as described below.  We make the entire table of results available in the supplementary material, noting that as of the time of this submission, we have included 147 model configurations; models for a small subset of experimental configurations (namely some of the high resolution SSD models) 
have yet to converge, so we have for now omitted them from analysis.

\subsection{Analyses}

\paragraph{Accuracy vs time}

Figure~\ref{fig:map_v_gputime_by_meta_and_feat} is a scatterplot visualizing the mAP
of each of our model configurations,
with colors representing feature extractors, and marker shapes representing
meta-architecture.  Running time per image ranges from tens of
milliseconds to almost 1 second.   
Generally we observe that  R-FCN and SSD models are faster on average
while Faster R-CNN tends to lead to slower but more accurate models,
requiring at least 100 ms per image.  However, as we discuss below,
Faster R-CNN models can be just as fast if we limit the number of
regions proposed.  
We have also overlaid an imaginary ``optimality frontier''
representing points at which better accuracy can only be attained
within this family of detectors by sacrificing
speed.
In the following, we highlight some of the key points along
the optimality frontier as the best detectors to use and discuss the
effect of the various model configuration options in isolation.

\begin{figure*}[t!]
\begin{center}
  \includegraphics[width=0.6\linewidth]{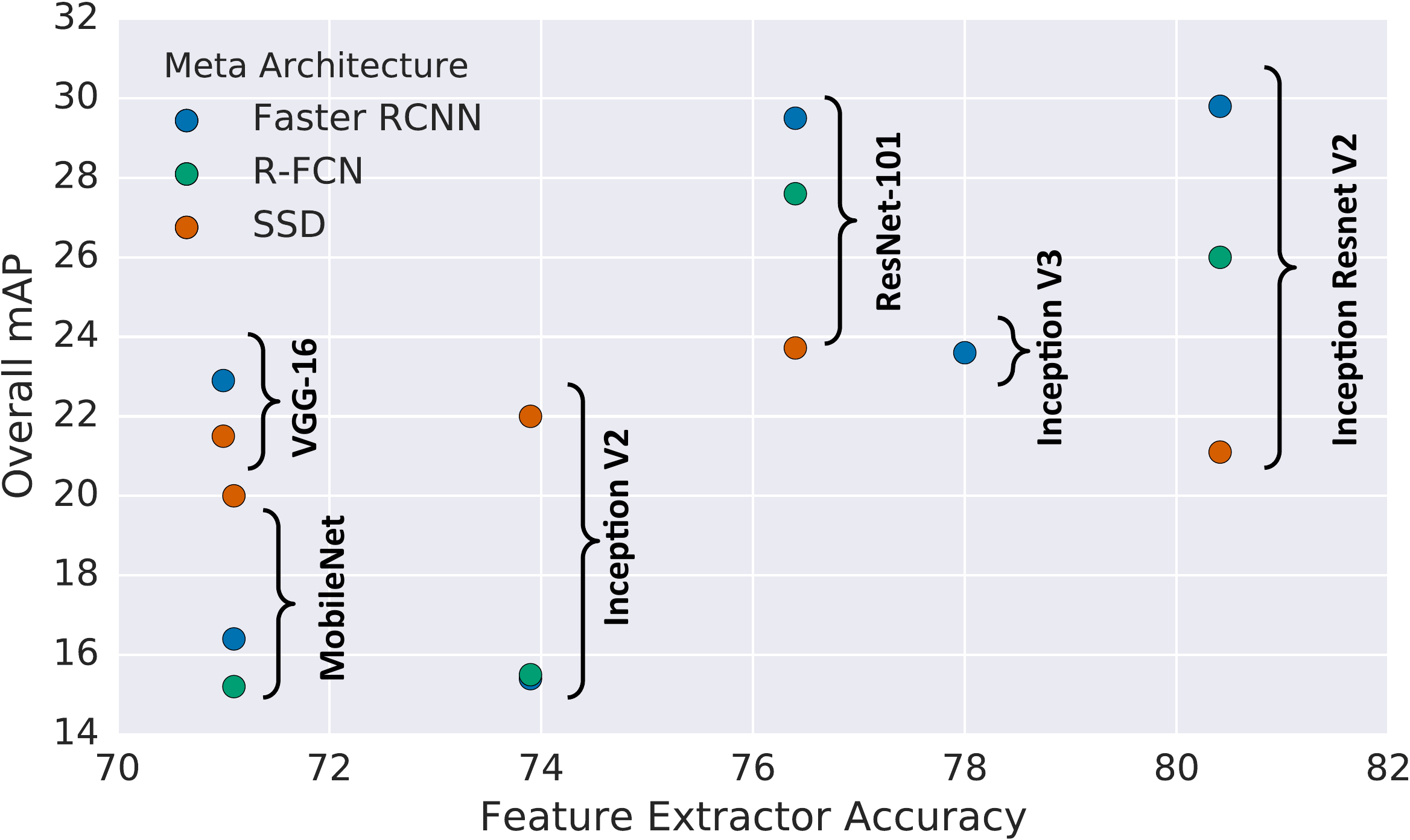}
\end{center}\vspace{-5mm}
\caption{
  \footnotesize
  Accuracy of detector (mAP on COCO) vs accuracy of feature extractor (as measured
  by top-1 accuracy on ImageNet-CLS). To avoid crowding the plot, we show only the low resolution models.
}
\label{fig:map_v_imagenet}
\end{figure*}

\begin{figure*}
\begin{center}
\includegraphics[
  width=.85\linewidth]{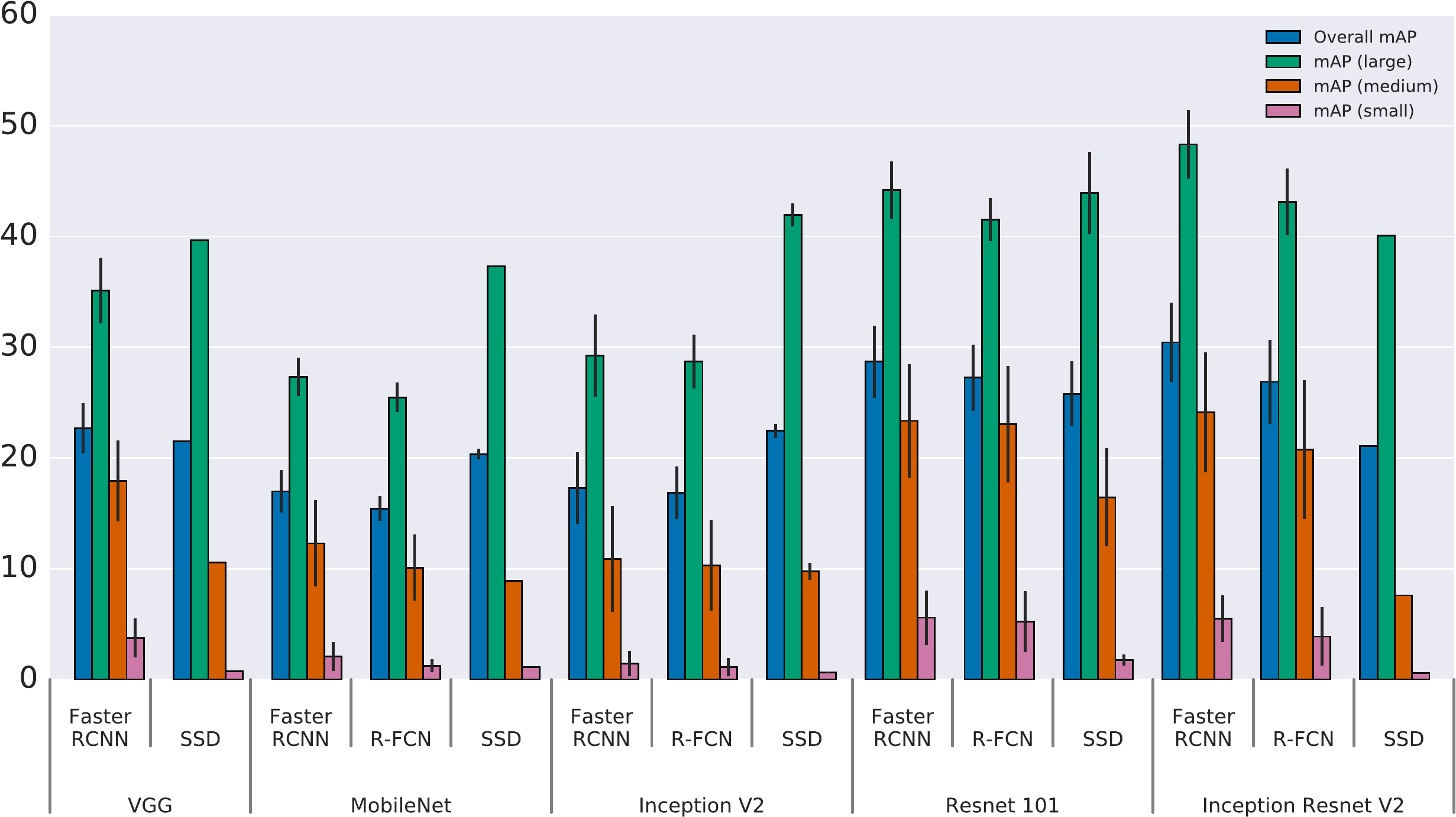}\vspace{-2mm}
\caption{
\footnotesize Accuracy stratified by  object size,  meta-architecture and feature extractor,
We fix the image resolution to 300.
}\vspace{-4mm}
\label{fig:map_by_object_size}
\end{center}
\end{figure*}

\paragraph{Critical points on the optimality frontier.}

{\em (Fastest: SSD w/MobileNet)}:
On the fastest end of this optimality frontier, we see that SSD models with Inception v2 and Mobilenet feature extractors are most accurate of the fastest models.  Note that if we ignore postprocessing costs, Mobilenet seems to be roughly twice as fast as Inception v2 while being slightly worse in accuracy.
{\em (Sweet Spot: R-FCN  w/Resnet or Faster R-CNN w/Resnet and only 50 proposals)}:
There is an ``elbow'' in the middle of the optimality frontier occupied by 
R-FCN models using Residual Network feature extractors which seem to strike
the best balance between speed and accuracy among our model configurations.
As we discuss below, Faster R-CNN w/Resnet models can attain similar speeds if we limit the number of proposals to 50.
{\em (Most Accurate: Faster R-CNN w/Inception Resnet at stride 8)}:
Finally Faster R-CNN with dense output Inception Resnet models attain
the best possible accuracy on our optimality frontier, achieving, to
our knowledge, the state-of-the-art single model performance.  However
these models are slow, requiring nearly a second of processing time.
The overall mAP numbers for these 5 models are shown in
Table~\ref{tab:testdev}.


\paragraph{The effect of the feature extractor.}

\begin{figure*}[t]
\begin{center}
\includegraphics[width=.9\linewidth]{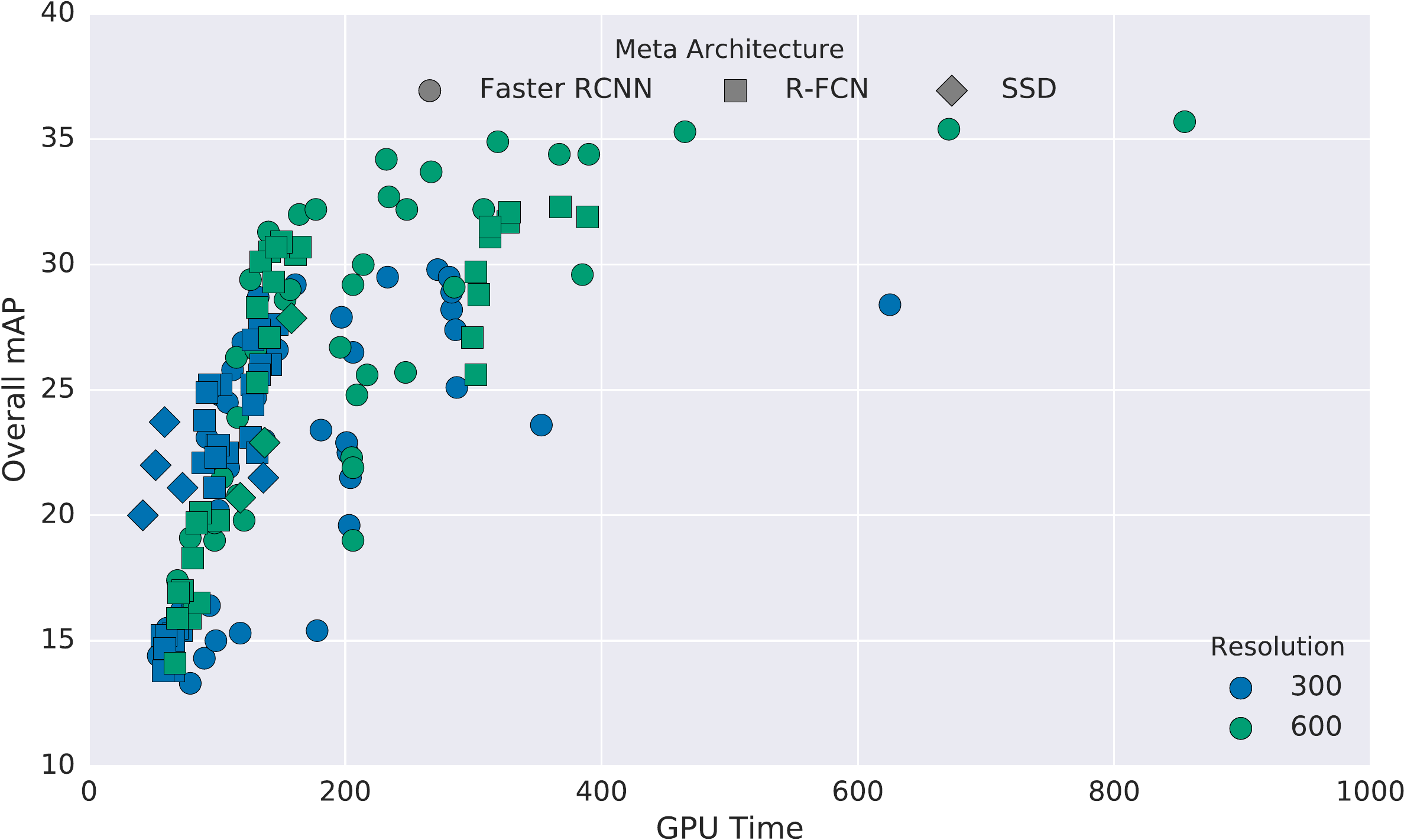}
\caption{
  \footnotesize Effect of image resolution.
}\vspace{-5mm}
\label{fig:map_v_gputime_by_size}
\end{center}
\end{figure*}

Intuitively, stronger performance on classification should be
positively correlated with stronger performance on COCO detection.   
To verify this, we investigate the relationship between overall mAP of different models and the Top-1 Imagenet classification accuracy attained by
the pretrained feature extractor used to initialize each model.
Figure~\ref{fig:map_v_imagenet} indicates that there is indeed an overall correlation between classification and detection performance.
However this correlation appears to  only be significant for Faster R-CNN and R-FCN while the performance of SSD appears to be less reliant on
its feature extractor's classification accuracy.  

\paragraph{The effect of object size.}
\vspace{-2mm}
Figure~\ref{fig:map_by_object_size} shows performance for different
models on different sizes of objects.
Not surprisingly, all methods do much better on large objects.
We also see that even though SSD models
typically have (very) poor performance on small objects, they are
competitive with Faster RCNN and R-FCN on large objects, even
outperforming these meta-architectures 
for the faster and more lightweight feature extractors.

\paragraph{The effect of image size.}

It has been observed by other authors that  input resolution can
significantly impact detection accuracy.  From our experiments, we
observe that decreasing resolution by a factor of two in both
dimensions consistently lowers accuracy (by $15.88\%$ on average) but
also reduces inference time by a relative factor of $27.4\%$ on
average. 

One reason for this effect is that high resolution inputs allow for
small objects to be resolved. Figure~\ref{fig:map_v_gputime_by_size}
compares detector performance on large objects against that on
small objects, confirms that high resolution models lead to
significantly better mAP results on small objects (by a factor of 2 in
many cases) and somewhat better mAP results on large objects as well.  
We also see that strong performance on small objects implies strong
performance on large objects in our models, (but not vice-versa as SSD
models do well on large objects but not small).

\paragraph{The effect of the number of proposals.}

\begin{figure*}
\begin{center}
\begin{subfigure}[t]{0.35\textwidth}
  \includegraphics[width=\linewidth]{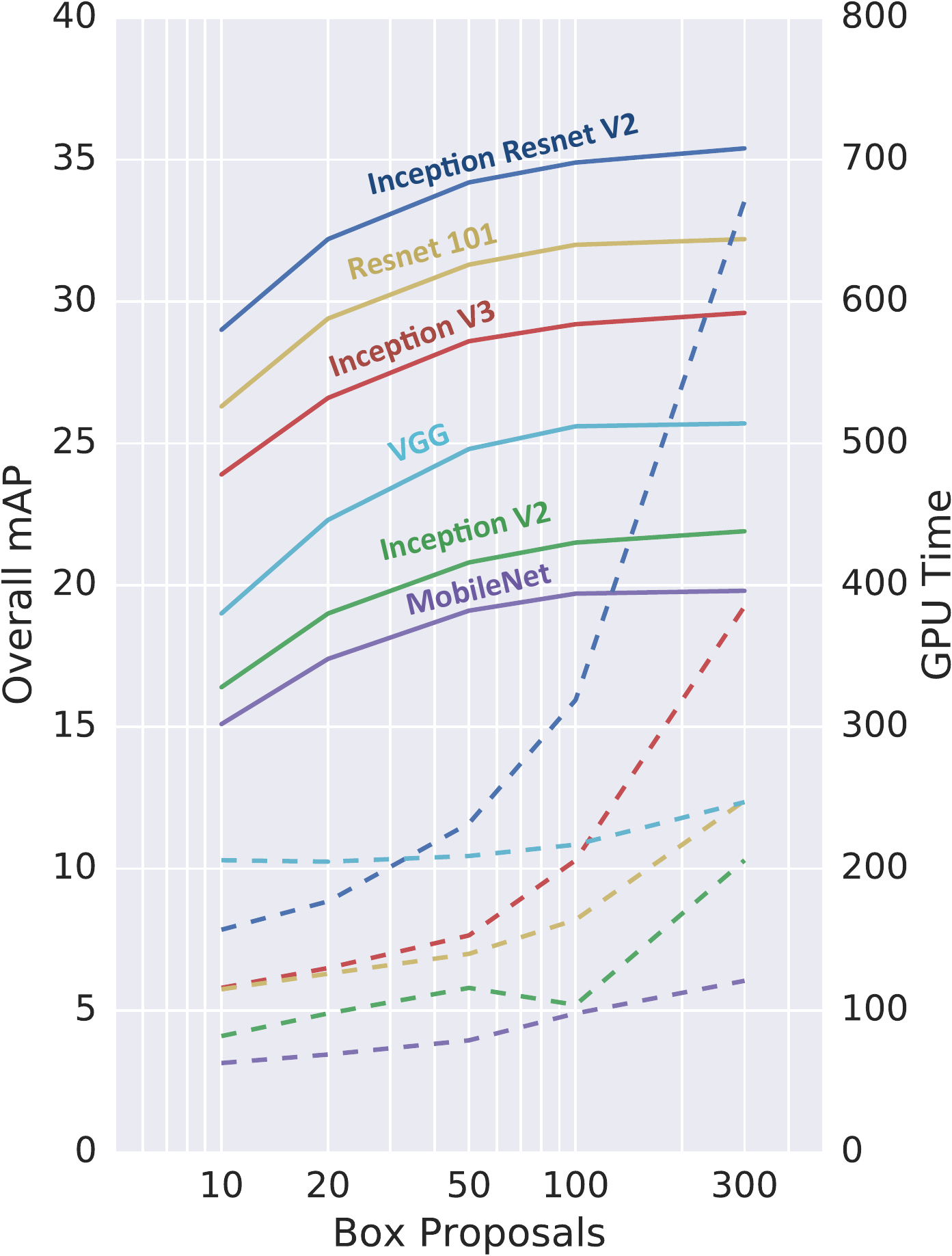}
  \caption{FRCNN}
\label{fig:faster_rcnn_proposals_tradeoff}
\end{subfigure}\qquad
\begin{subfigure}[t]{0.35\linewidth}
  \includegraphics[width=\linewidth]{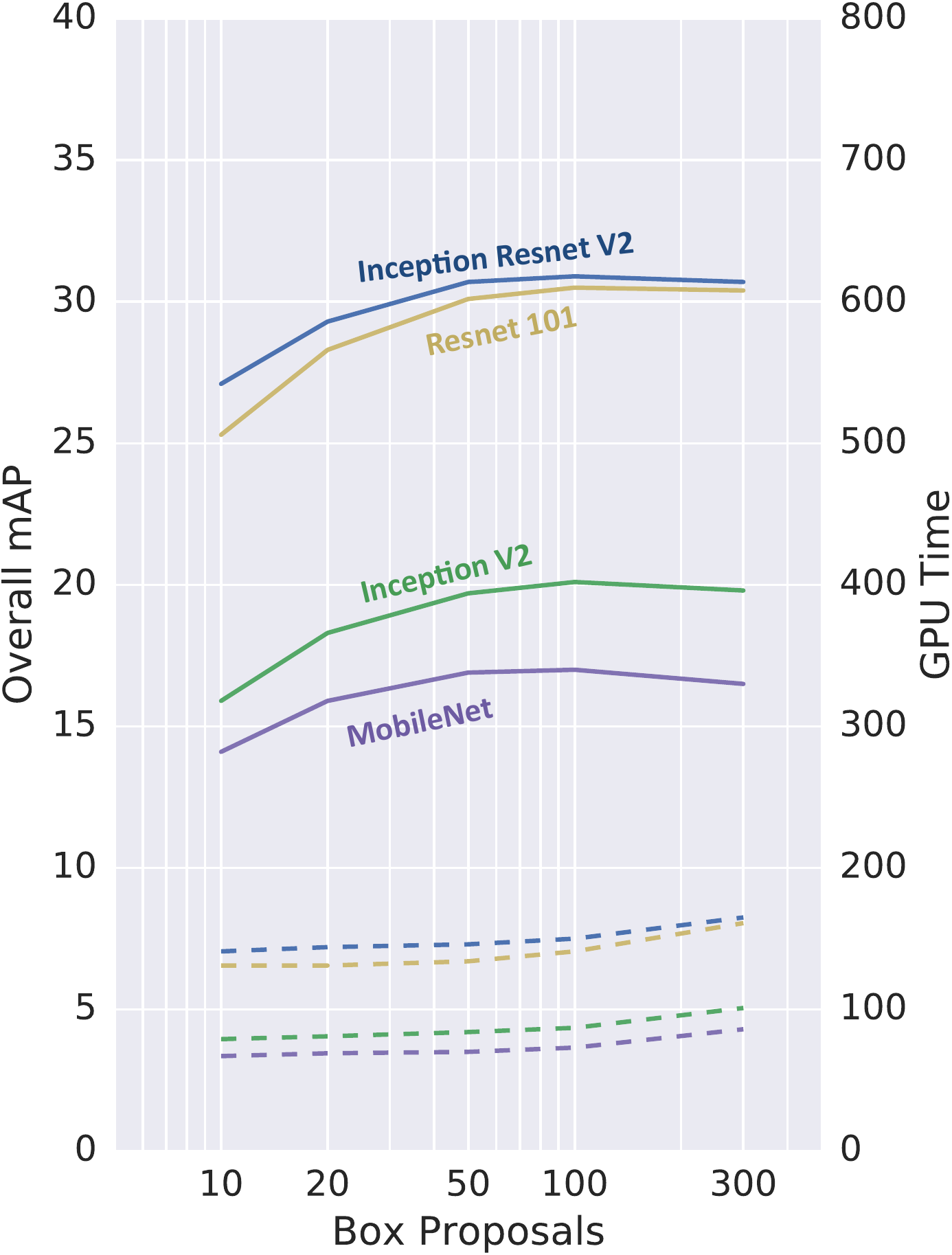}
  \caption{RFCN}
\label{fig:rfcn_proposals_tradeoff}
\end{subfigure}
\end{center}
\caption{
\footnotesize  
Effect of proposing increasing number of regions
on mAP accuracy (solid lines) and
GPU inference time (dotted).
Surprisingly, for Faster R-CNN with Inception
Resnet, we obtain 96\% of the accuracy of using 300 proposals by using
only 50 proposals, which reduces running time by a factor of 3. 
}
\end{figure*}

\begin{figure*}[t!]
\begin{center}
\includegraphics[width=.85\linewidth]{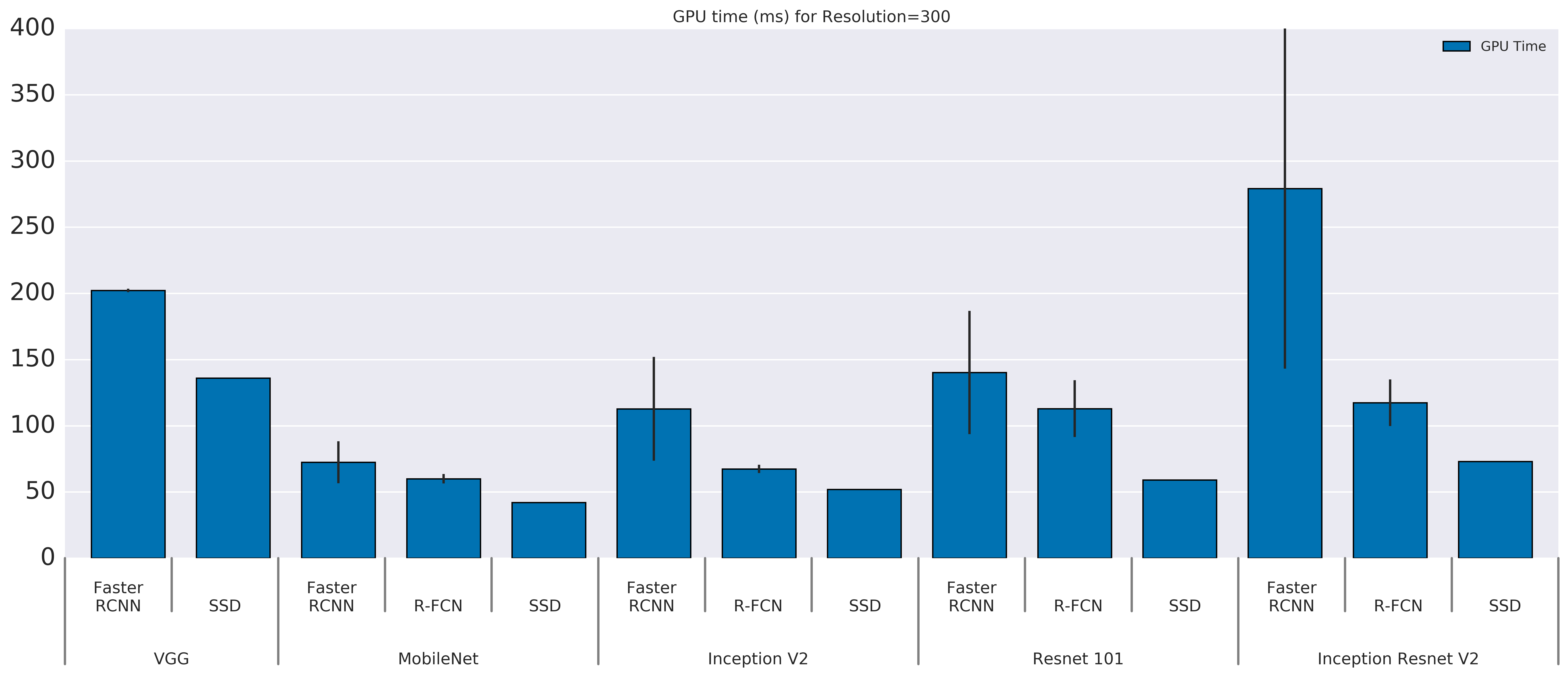}\vspace{-4mm}
\caption{
\footnotesize GPU time (milliseconds) for each model,
for image resolution of 300.
}\vspace{-4mm}
\label{fig:bar_gpu_time}
\end{center}
\end{figure*}

For Faster R-CNN and R-FCN, we can adjust the number of proposals
computed by the region proposal network.  The authors in both papers
use 300 boxes, however, our experiments suggest that this number can
be significantly reduced without harming mAP (by much).  In some
feature extractors where the ``box classifier'' portion of Faster
R-CNN is expensive, this can lead to significant computational
savings.  Figure~\ref{fig:faster_rcnn_proposals_tradeoff} visualizes
this trade-off curve for Faster R-CNN models with high resolution
inputs for different feature extractors.   
We see that Inception Resnet, which has 35.4\% mAP with 300 proposals
can still have surprisingly high accuracy (29\% mAP) with only 10
proposals.  The sweet spot is probably at 50 proposals, where we are
able to obtain 96\% of the accuracy of using 300 proposals while
reducing running time by a factor of 3.  While the computational
savings are most pronounced for Inception Resnet, we see that similar
tradeoffs hold for all feature extractors. 

Figure~\ref{fig:rfcn_proposals_tradeoff} visualizes the same trade-off
curves for R-FCN models and shows that the computational savings from
using fewer proposals in the R-FCN setting are minimal --- this is not
surprising as the box classifier (the expensive part) is only run once
per image.  We see in fact that at 100 proposals, the speed and
accuracy for Faster R-CNN models with ResNet becomes roughly
comparable to that of equivalent R-FCN models which use 300 proposals
in both mAP and GPU speed. 

\paragraph{FLOPs analysis.}\vspace{-3mm}

\begin{figure*}[t!]
{\footnotesize 
  \begin{center}
    \begin{subfigure}[t]{0.48\linewidth}
\includegraphics[width=\linewidth]{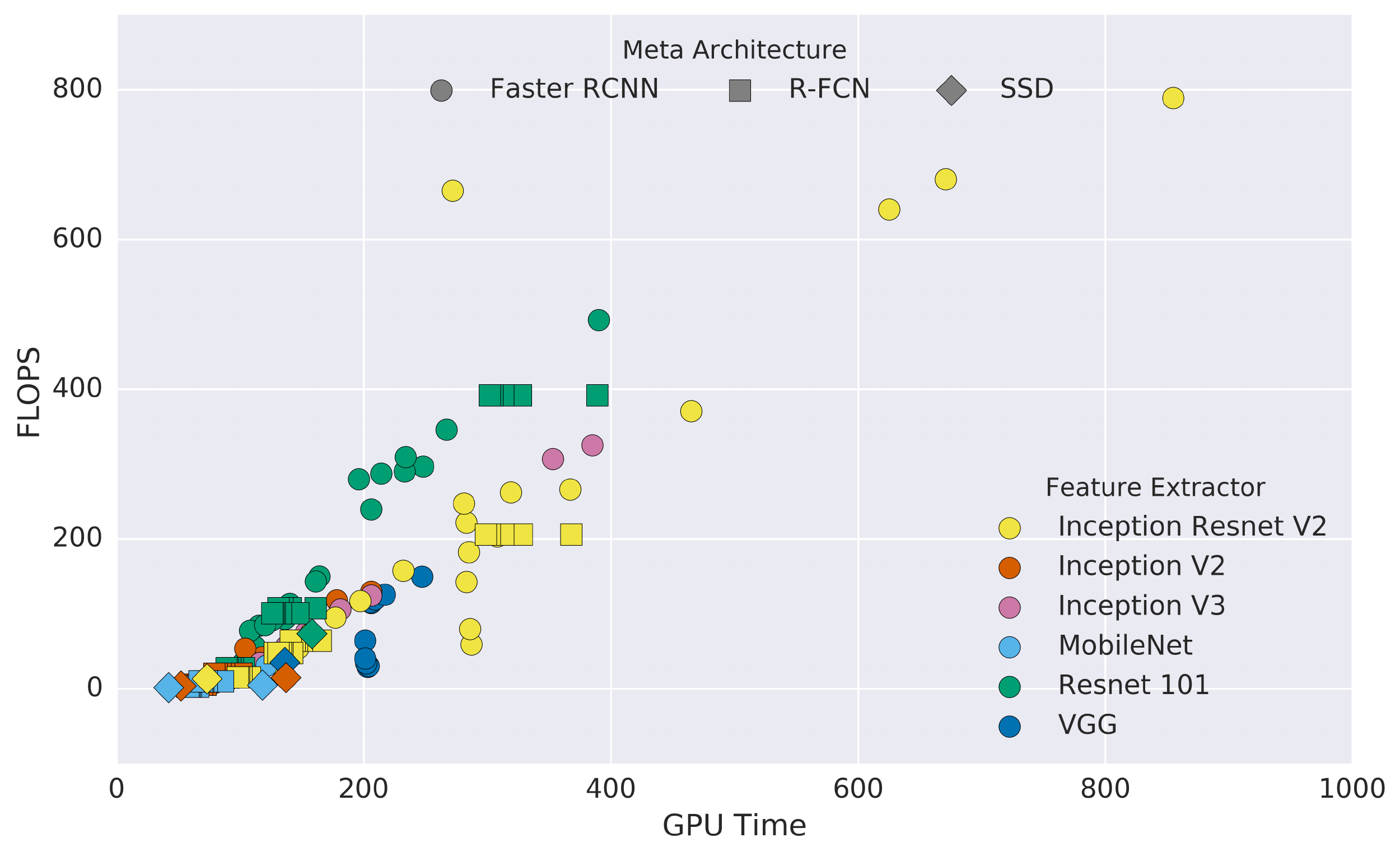}
\caption{GPU.}
\label{fig:flops_v_gputime}
    \end{subfigure}
        \begin{subfigure}[t]{0.48\linewidth}
\includegraphics[width=\linewidth]{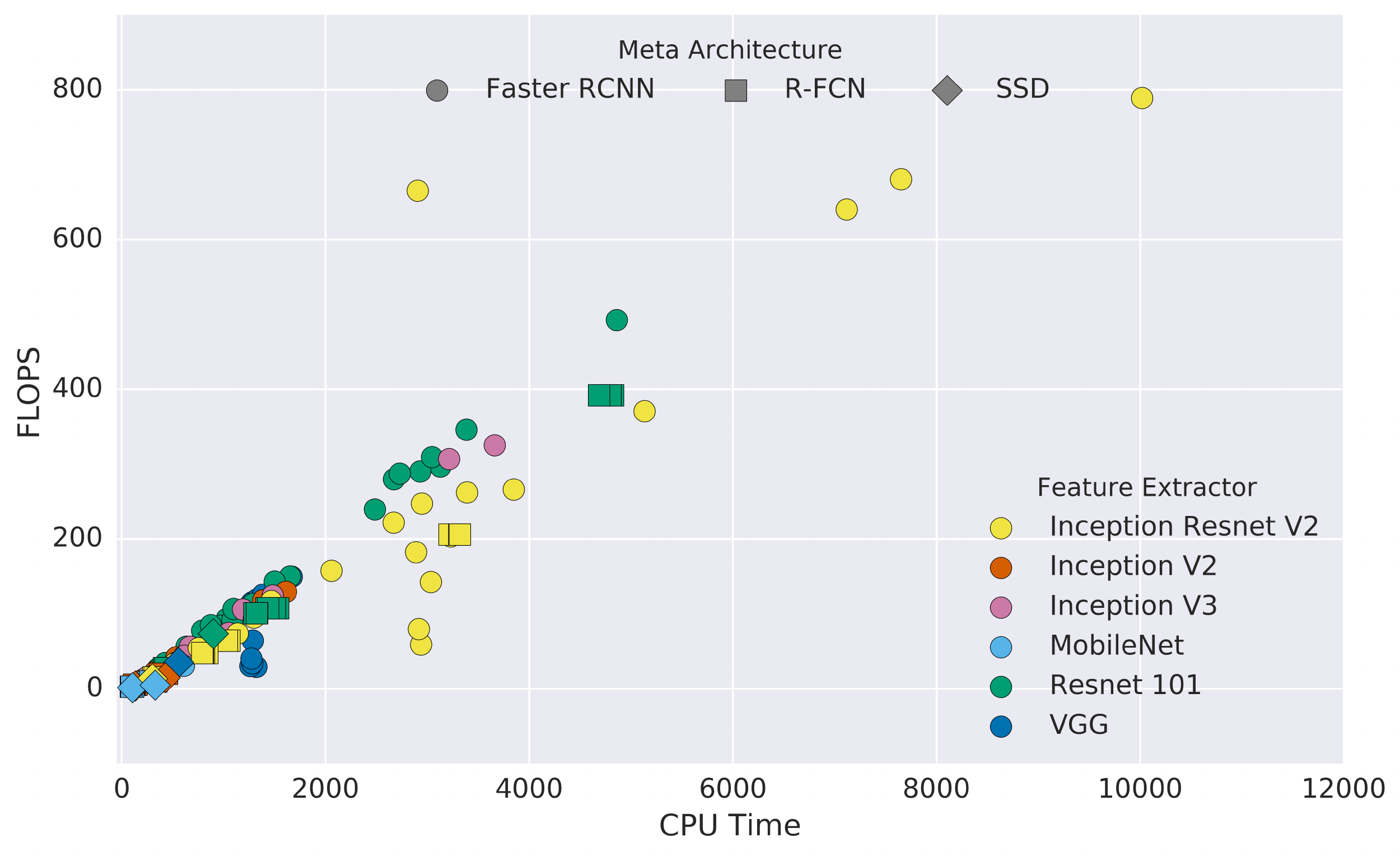}
\caption{CPU.}
\label{fig:flops_v_cputime}
\end{subfigure}\vspace{-5mm}
  \end{center}
  \caption{\footnotesize FLOPS vs time.}
  \label{fig:flops}
  }
\end{figure*}

\begin{figure*}[t!]
\begin{center}
\includegraphics[width=.85\linewidth]{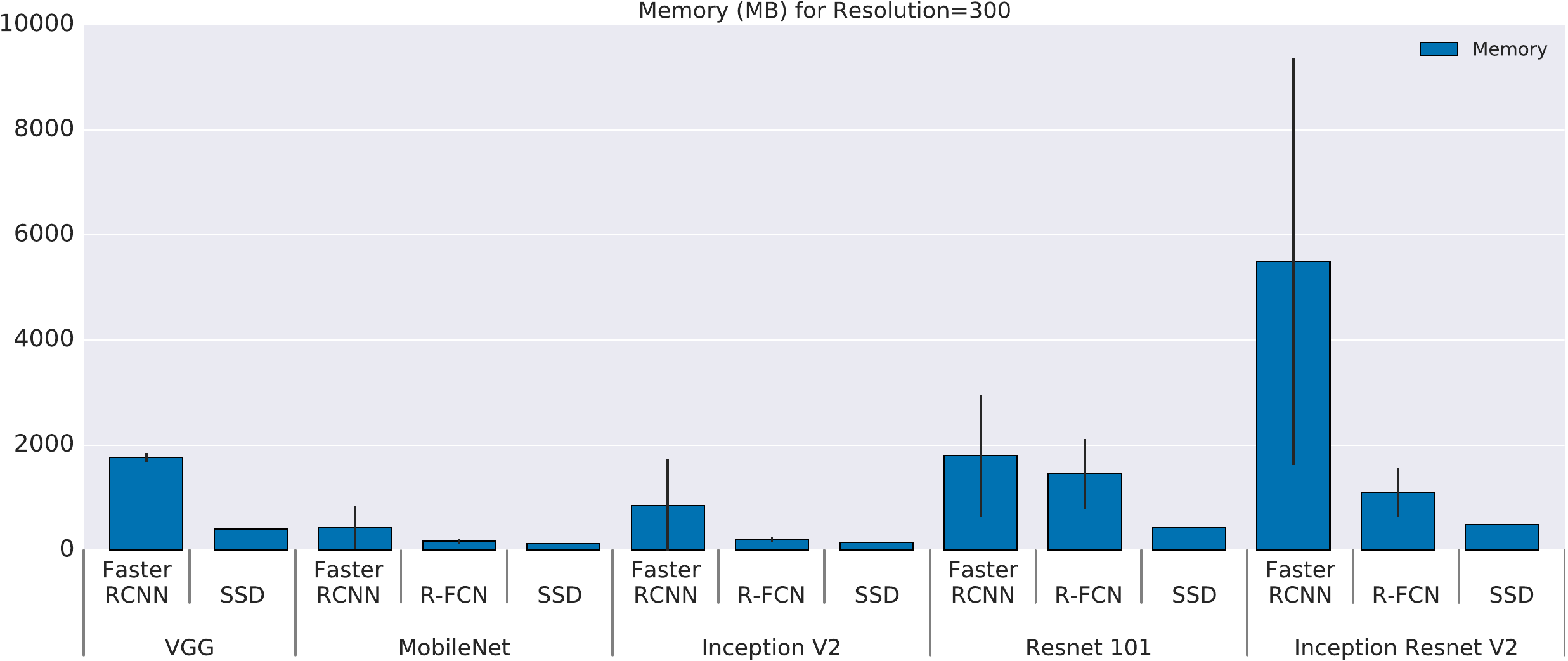}
\caption{
\footnotesize Memory (Mb) usage for each model.  Note that we measure total memory usage
rather than peak memory usage.  Moreover, we include all data points
corresponding to the low-resolution models here.  The error bars
reflect 
variance in memory usage by using different numbers of proposals for
the Faster R-CNN and R-FCN models (which leads to the seemingly 
considerable variance in the Faster-RCNN with Inception Resnet bar).
}
\label{fig:memory_barplot}
\end{center}
\end{figure*}

Figure~\ref{fig:bar_gpu_time} plots the GPU time for each model
combination.
However, this is very platform dependent.
Counting FLOPs (multiply-adds)
gives us a platform independent measure of computation,
which may or may not be linear with respect to actual running times
due to a number of issues such as caching, I/O, hardware optimization etc,  

Figures~\ref{fig:flops_v_gputime} and~\ref{fig:flops_v_cputime} plot
the FLOP count against observed wallclock times on the GPU and CPU respectively.
Interestingly,  we observe in the GPU plot (Figure~\ref{fig:flops_v_gputime})
that each model has a different average ratio of flops to observed
running time in milliseconds.  For denser block models such as Resnet
101, FLOPs/GPU time is typically greater than 1, perhaps due to
efficiency in caching. For Inception and Mobilenet models, this ratio
is typically less than 1 --- we conjecture that this could be that
factorization reduces FLOPs, but adds more overhead in memory I/O or
potentially that current GPU instructions (cuDNN) are more optimized
for dense convolution.

\paragraph{Memory analysis.}

\begin{figure*}[t!]
{\footnotesize 
  \begin{center}
    \begin{subfigure}{0.48\linewidth}
\includegraphics[width=\linewidth]{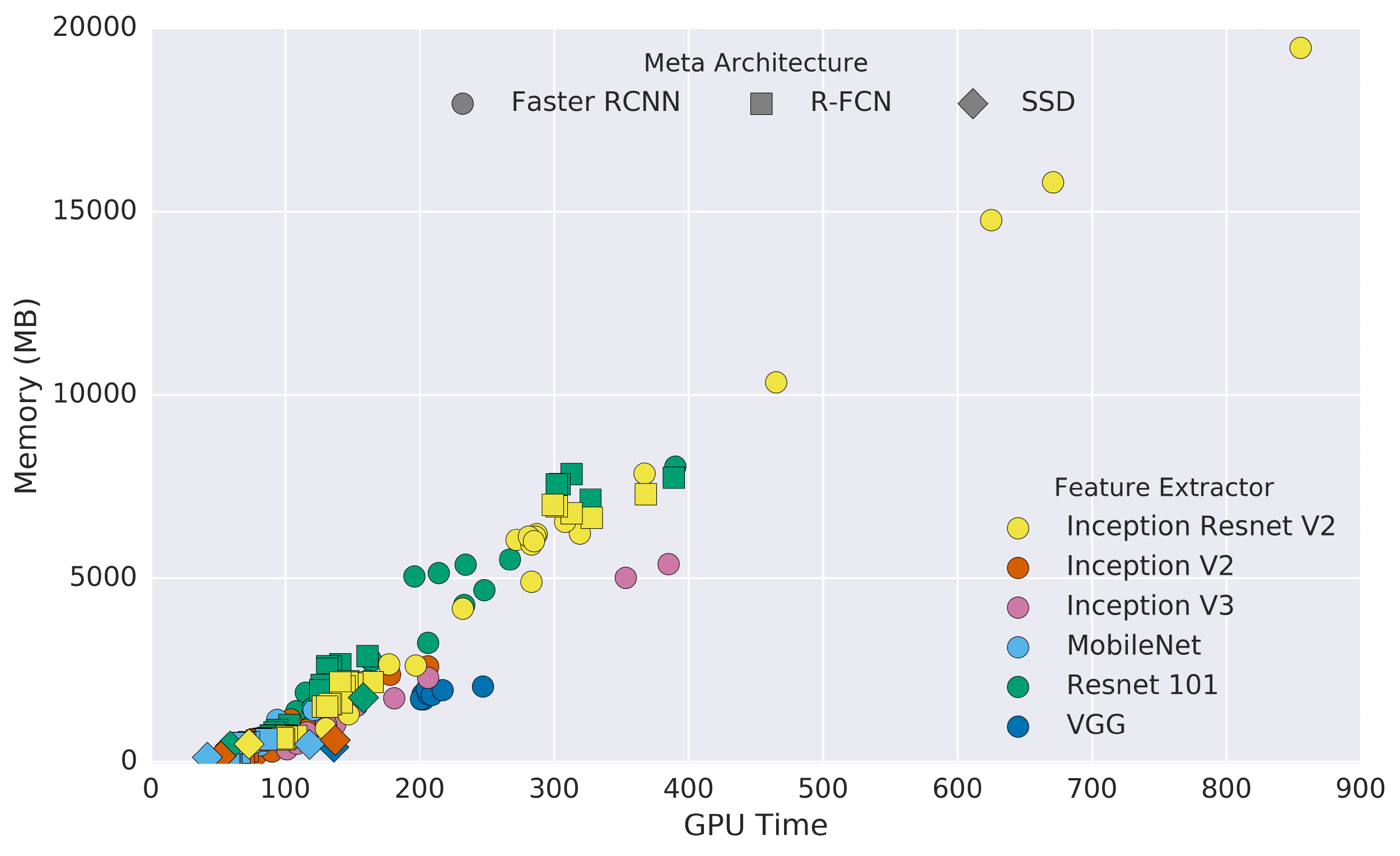}
\caption{GPU}.
\label{fig:memory_v_gputime}
\end{subfigure}\qquad
    \begin{subfigure}{0.48\linewidth}
\includegraphics[width=\linewidth]{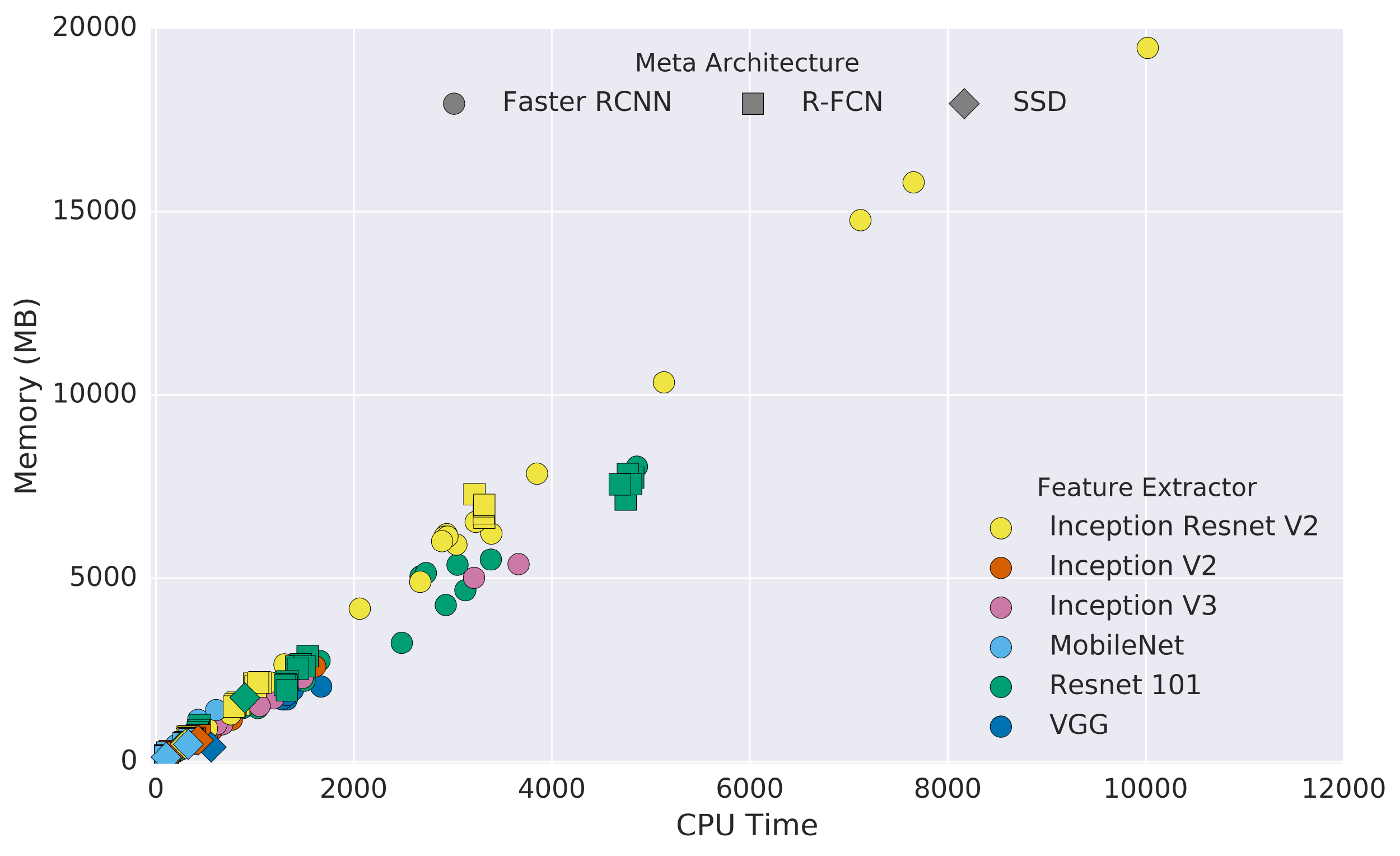}
\caption{CPU}.
\label{fig:memory_v_cputime}
\end{subfigure}\vspace{-7mm}
  \end{center}
  \caption{\footnotesize Memory (Mb) vs time.}
  \label{fig:memory}
  }
\end{figure*}

For memory benchmarking, we measure total usage rather than peak
usage.  Figures~\ref{fig:memory_v_gputime},~\ref{fig:memory_v_cputime}
plot memory usage against GPU and CPU wallclock times.  Overall, we
observe high correlation with running time with larger and more
powerful feature extractors requiring much more
memory. Figure~\ref{fig:memory_barplot} plots some of the same
information in more detail, drilling down by meta-architecture and
feature extractor selection.  As with speed, Mobilenet is again the
cheapest, requiring less than 1Gb (total) memory in almost all
settings.

\paragraph{Good localization at .75 IOU means good localization at all IOU thresholds.}

\begin{figure*}
\begin{center}
\includegraphics[width=0.45\linewidth]{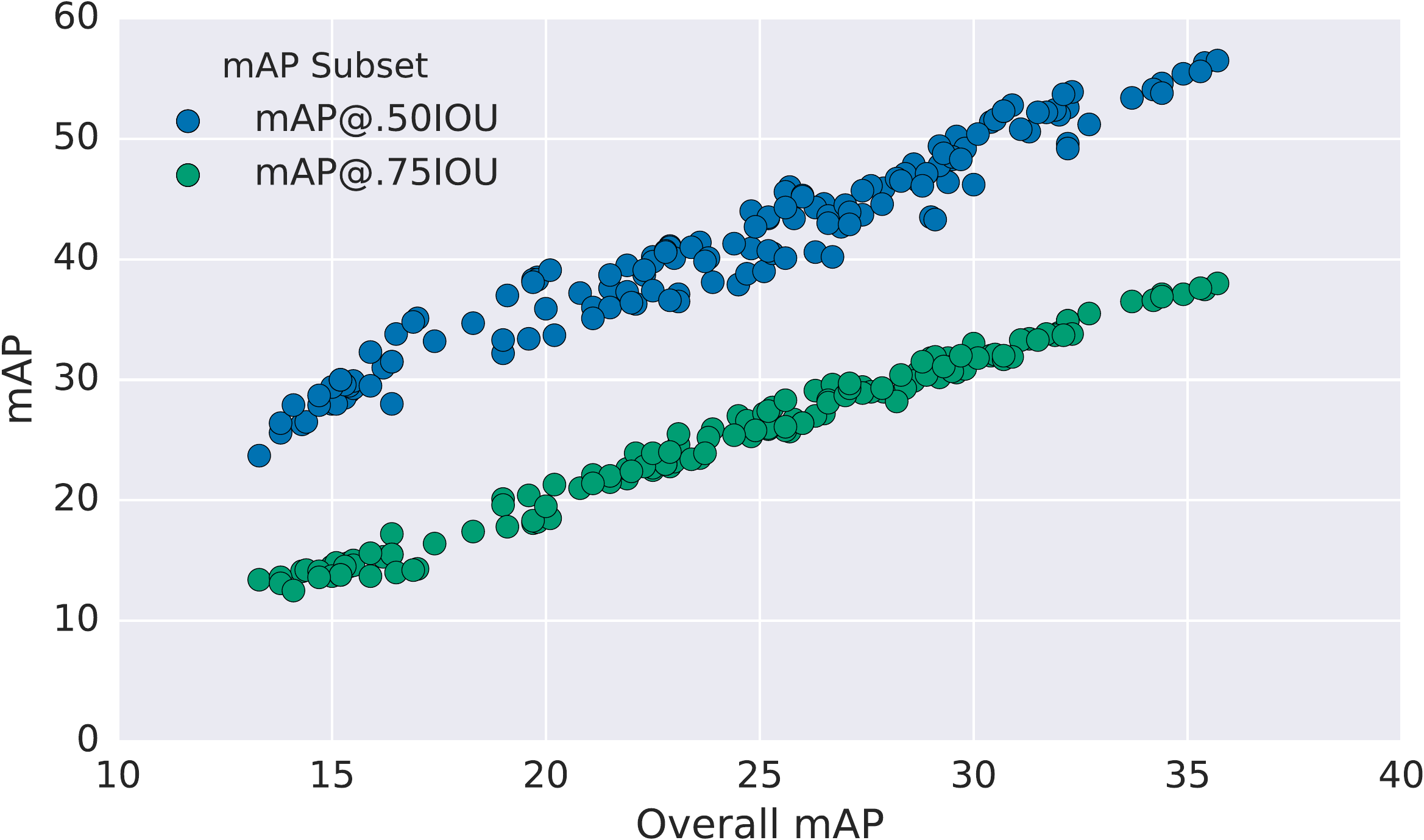}
\end{center}\vspace{-4mm}
\caption{
\footnotesize Overall COCO mAP (@[.5:.95]) for all experiments plotted against
corresponding mAP@.50IOU and mAP@.75IOU.  It is unsurprising that these numbers
are correlated, but it is interesting that they are almost perfectly correlated so for these models, it is never the case that a model has strong 
performance at 50\% IOU but weak performance at 75\% IOU.
}
\label{fig:map_50_v_75}
\end{figure*}

While slicing the data by object size leads to interesting insights,
it is also worth nothing that slicing data by IOU threshold does not
give much additional information.  Figure~\ref{fig:map_50_v_75} shows
in fact that both  mAP@.5 and mAP@.75 performances are almost
perfectly linearly correlated with mAP@[.5:.95].  Thus detectors that
have poor performance at the higher IOU thresholds always also show
poor performance at the lower IOU thresholds. 
This being said, we also observe that mAP@.75 is slightly more tightly
correlated with mAP@[.5:.95] (with $R^2>.99$), so if we were to
replace the standard COCO metric with mAP at a single IOU threshold,
we would likely choose IOU=.75.

\begin{table*}[t!]
\begin{center}
{\footnotesize
\begin{tabular}{c|c|c|c|c|c|c|c|c|c|c}
\, & AP & AP@.50IOU & AP@.75IOU & AP$_{small}$ & AP$_{med}$ & AP$_{large}$ & AR@100 & AR$_{small}$ & AR$_{med}$ & AR$_{large}$ \\
\hline
Ours & \bf 0.413 & \bf 0.62 & \bf 0.45 & \bf 0.231 & \bf 0.436 & \bf 0.547 & \bf 0.604 & \bf 0.424 & \bf 0.641 & \bf 0.748 \\
MSRA2015 & 0.371 & 0.588 & 0.398 & 0.173 & 0.415 & 0.525 & 0.489 & 0.267 & 0.552 & 0.679 \\
Trimps-Soushen & 0.359 & 0.58 & 0.383 & 0.158 & 0.407 & 0.509 & 0.497 & 0.269 & 0.557 & 0.683 \\
\end{tabular}\vspace{-4mm}
}
\end{center}
\caption{
\footnotesize Performance on the  2016 COCO test-challenge dataset.
AP and AR refer to (mean) average precision and average recall respectively.
Our model achieves a relative improvement of nearly 60\% on small objects recall over the previous state-of-the-art COCO detector.
}
\label{tab:cocoresults}
\end{table*}

\subsection{State-of-the-art detection on COCO}

Finally, we briefly describe how we ensembled some of our models to
achieve the current state of the art performance on the 2016 COCO
object detection challenge.  Our model attains 41.3\% mAP@[.5, .95] on
the COCO test set and is an ensemble of five Faster R-CNN models based
on Resnet and Inception Resnet feature extractors.
This outperforms the
previous best result (37.1\% mAP@[.5, .95]) by MSRA, which used an
ensemble of three  
Resnet-101 models~\cite{he2015deep}.
Table~\ref{tab:cocoresults} summarizes the performance of our model
and highlights how our model has improved on the state-of-the-art
across all COCO metrics.  Most notably, our model achieves a relative
improvement of nearly 60\% on small object recall over the previous
best result. 
Even though this ensemble with state-of-the-art numbers could be
viewed as an extreme point on the speed/accuracy tradeoff curves
(requires $\sim$50 end-to-end network evaluations per image), we have
chosen to present this model in isolation since it is not comparable
to the ``single model'' results that we focused on in the rest of the paper.

\begin{table*}[t!]
\begin{center}
{\footnotesize
\begin{tabular}{c|c|c|c|c}
AP & Feature Extractor & Output stride & loss ratio & Location loss function \\
\hline
32.93 & Resnet 101 & 8 & 3:1 & SmoothL1 \\
33.3 & Resnet 101 & 8 & 1:1 & SmoothL1 \\
34.75 & Inception Resnet (v2) & 16 & 1:1 & SmoothL1 \\
35.0 & Inception Resnet (v2) & 16 & 2:1 & SmoothL1 \\
35.64 & Inception Resnet (v2) & 8 & 1:1 & SmoothL1 +  IOU 
\end{tabular}
}
\end{center}
\caption{
\footnotesize 
Summary of single models that were automatically selected to be part
of the diverse ensemble.
Loss ratio refers to the multipliers
$\alpha, \beta$ for location and classification losses, respectively.
}
\label{tab:models_for_ensemble}
\end{table*}

\begin{table*}[t!]
\begin{center}
{\footnotesize
\begin{tabular}{c|c|c|c|c|c|c}
\, & AP & AP@.50IOU & AP@.75IOU & AP$_{small}$ & AP$_{med}$ & AP$_{large}$ \\
\hline
Faster RCNN with Inception Resnet (v2) & 0.347  & 0.555 & 0.367 &0.135 & 0.381  & 0.52 \\
Hand selected Faster RCNN ensemble w/multicrop & 0.41 & 0.617 & 0.449 & 0.236 & 0.43 & 0.542 \\
Diverse Faster RCNN ensemble w/multicrop & 0.416 & 0.619 & 0.454 & 0.239 & 0.435 & 0.549
\end{tabular}
}
\caption{
\footnotesize Effects of ensembling and multicrop inference.  Numbers reported on COCO test-dev dataset.
Second row (hand selected ensemble) consists of 6 Faster RCNN models
with 3 Resnet 101 (v1) and 3 Inception Resnet (v2) and the third row
(diverse ensemble) is described in detail in
Table~\ref{tab:models_for_ensemble}. 
}
\label{tab:improvement_over_single_model}
\end{center}
\end{table*}

To construct our ensemble, we selected a set of five models from our
collection of Faster R-CNN models.  Each of the models was based on
Resnet and Inception Resnet feature extractors with varying output
stride configurations, retrained using variations on the loss
functions, and different random orderings of the training data.
Models were selected greedily using their performance on a held-out
validation set.  However, in order to take advantage of models with
complementary strengths, we also explicitly encourage diversity by
pruning away models that are too similar to previously selected
models (c.f., \cite{Lee2015diverse}).
To do this, we computed the vector of average precision
results across each COCO category for each model and declared two
models to be too similar if their category-wise AP vectors had cosine
distance greater than some threshold.

Table~\ref{tab:models_for_ensemble} summarizes the final selected
model specifications as well as their individual performance on COCO
as single models.\footnote{Note that these numbers were computed on a
  held-out validation set and are not strictly comparable to the
  official COCO test-dev data results (though they are expected to be
  very close).}  Ensembling these five models using the procedure
described in~\cite{he2015deep} (Appendix A) and using multicrop
inference then yielded our final model.  Note that we do not use
multiscale training, horizontal flipping, box refinement, box voting,
or global context which are sometimes used in the literature.
Table~\ref{tab:improvement_over_single_model} compares a single
model's performance against two ways of ensembling, and shows that (1)
encouraging for diversity did help against a hand selected ensemble,
and (2) ensembling and multicrop were responsible for almost 7 points
of improvement over a single model.

\input{examples}

%% file: examples.tex
\subsection{Example detections}
\label{sec:exampls}

In Figures~\ref{fig:export-image-10} to \ref{fig:export-image-1089}
we visualize detections on images from the COCO dataset, showing side-by-side comparisons of five
of the detectors that lie on the ``optimality frontier'' of the speed-accuracy
trade-off plot.  To visualize, we select detections with score greater
than a  threshold and plot the top 20 detections in each image.  We
use a threshold of .5 for Faster R-CNN and R-FCN and .3 for SSD.
These thresholds were hand-tuned for (subjective) visual
attractiveness and not using  rigorous criteria so we caution viewers
from reading too much into the tea leaves from these visualizations.
This being said, we see that across our examples, all of the detectors
perform reasonably well on large objects --- SSD only shows its
weakness on small objects, missing some of the smaller kites and
people in the first image as well as the smaller cups and bottles on
the dining table in the last image.  

\newcommand{\figdir}{Figures/cocovis}

\newcommand{\addfig}[2] 
{
    \begin{figure*}
      \centering
        \begin{subfigure}[t]{0.4\textwidth}
        \centering\includegraphics[height=#2]{\figdir/ssd_mobilenet/#1.jpg}
      \caption{SSD+Mobilenet, lowres}
        \end{subfigure}
        \begin{subfigure}[t]{0.4\textwidth}
        \centering\includegraphics[height=#2]{\figdir/ssd_inceptionv2/#1.jpg}
      \caption{SSD+InceptionV2, lowres}
        \end{subfigure}
        \begin{subfigure}[t]{0.4\textwidth}
        \centering\includegraphics[height=#2]{\figdir/fasterrcnn_resnet101_100/#1.jpg}
      \caption{FRCNN+Resnet101, 100 proposals}
        \end{subfigure}
        \begin{subfigure}[t]{0.4\textwidth}
        \centering\includegraphics[height=#2]{\figdir/rfcn_resnet101_300/#1.jpg}
      \caption{RFCN+Resnet10, 300 proposals}
        \end{subfigure}
        \begin{subfigure}[t]{0.4\textwidth}
        \centering\includegraphics[height=#2]{\figdir/fasterrcnn_inceptionresnetv2_300/#1.jpg}
      \caption{FRCNN+IncResnetV2, 300 proposals}
        \end{subfigure}
        \caption{Example detections from 5 different models.}
        \label{fig:#1}
    \end{figure*}
}

\addfig{export-image-10}{45mm}  
\addfig{export-image-384}{50mm}
\addfig{export-image-408}{45mm} 
\addfig{export-image-933}{50mm}
\addfig{export-image-941}{45mm} 
\addfig{export-image-1089}{45mm} 

%% file: concl.tex
\section{Conclusion}
\label{sec:concl}

We have performed an experimental comparison of some of the main
aspects that influence the speed and accuracy of modern object
detectors. We hope this will help practitioners choose an appropriate
method when deploying object detection in the real world.
We have also identified some new techniques for improving speed
without sacrificing much accuracy, such as using many fewer proposals
than is usual for Faster R-CNN.

\eat{ 

Some limitations of this study
+ limited to the meta-architectures that we have studied
+ some of our conclusions are likely to be closely tied to specific choices made for selecting anchors.

The factors that pay off the most in improving accuracy seems to be...
}

%% file: extra_analyses.tex
\section{Extra Plots and Analyses}\label{sec:extra_analyses}

\begin{figure*}[t!]
\begin{center}
\includegraphics[
	width=.9\linewidth]{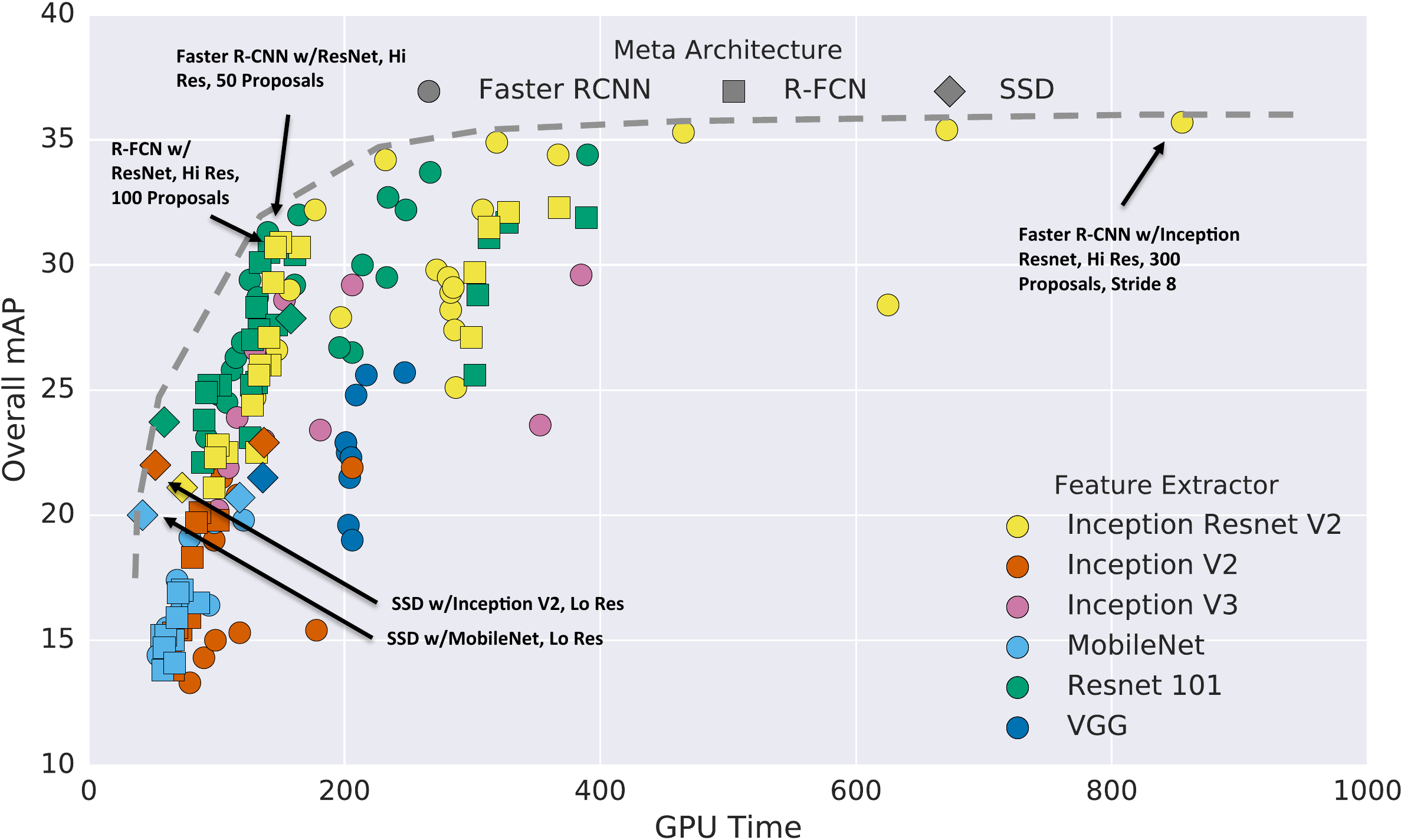}
\caption{
\footnotesize mAP vs gpu wallclock time 
	with marker shapes indicating 
	 meta-architecture
	and colors indicating  feature extractor. 
Each (meta-architecture, feature extractor) pair can correspond to multiple points on this plot due to changing input sizes, stride, etc.
}
\label{fig:map_v_gputime_by_meta_and_feat}
\end{center}\vspace{8mm}
\end{figure*}

\begin{figure*}[t]
\begin{center}
\includegraphics[
	width=.9\linewidth]{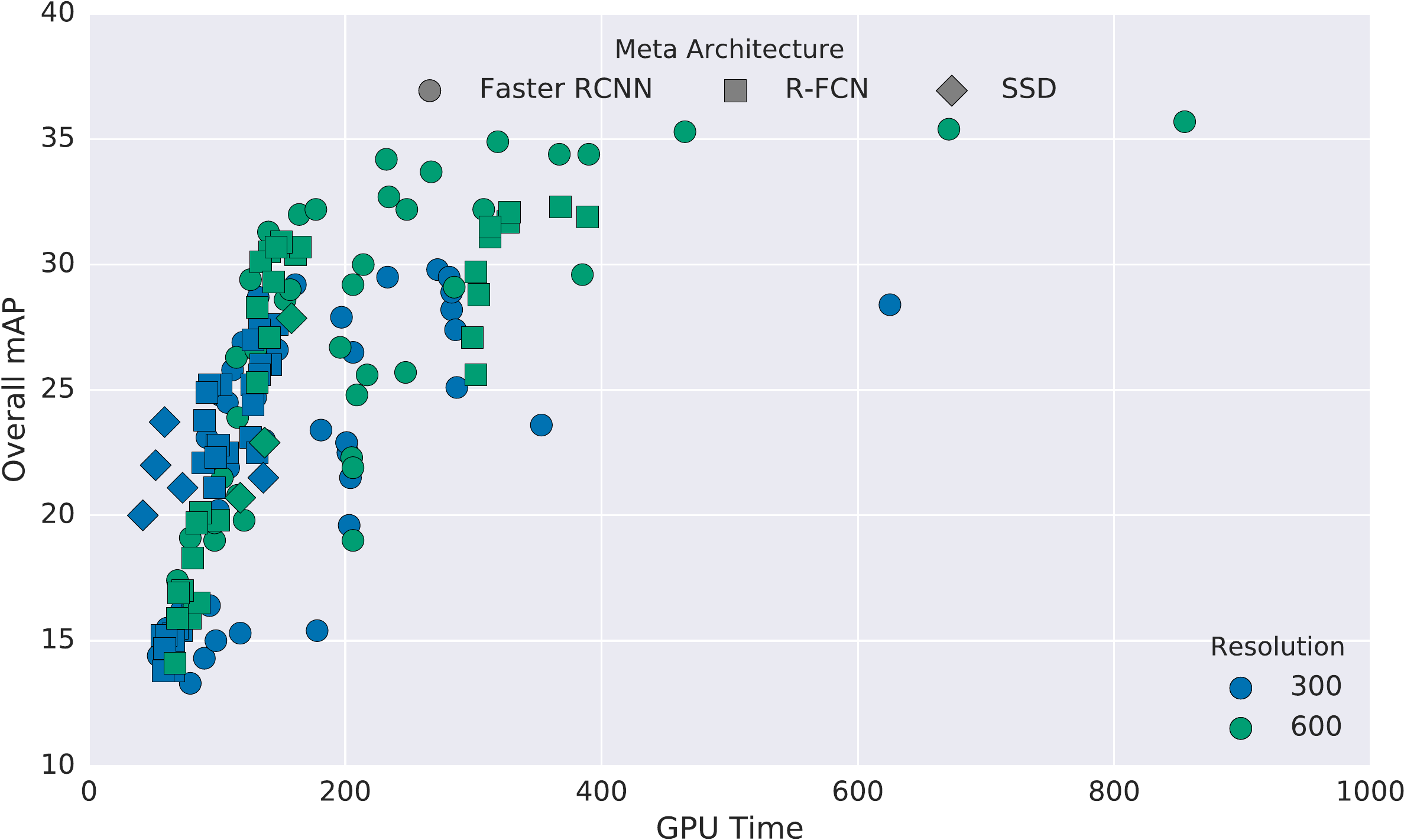}
\caption{
\footnotesize mAP vs gpu wallclock time 
	with marker shapes indicating 
	 meta-architecture
	and colors indicating image resolution. 
}
\label{fig:map_v_gputime_by_size}
\end{center}
\end{figure*}

\paragraph{mAP vs. GPU Wallclock time.}
Figures~\ref{fig:map_v_gputime_by_meta_and_feat}
and ~\ref{fig:map_v_gputime_by_size} are larger versions of our most important plots from the main paper.  Here we vary the marker shapes
to indicate meta-architecture and the colors to indicate feature extractor (in 
Figure~\ref{fig:map_v_gputime_by_meta_and_feat}) 
and image resolution (in Figure~\ref{fig:map_v_gputime_by_size}).

\begin{figure*}[t!]
\begin{center}
\includegraphics[
	width=.9\linewidth]{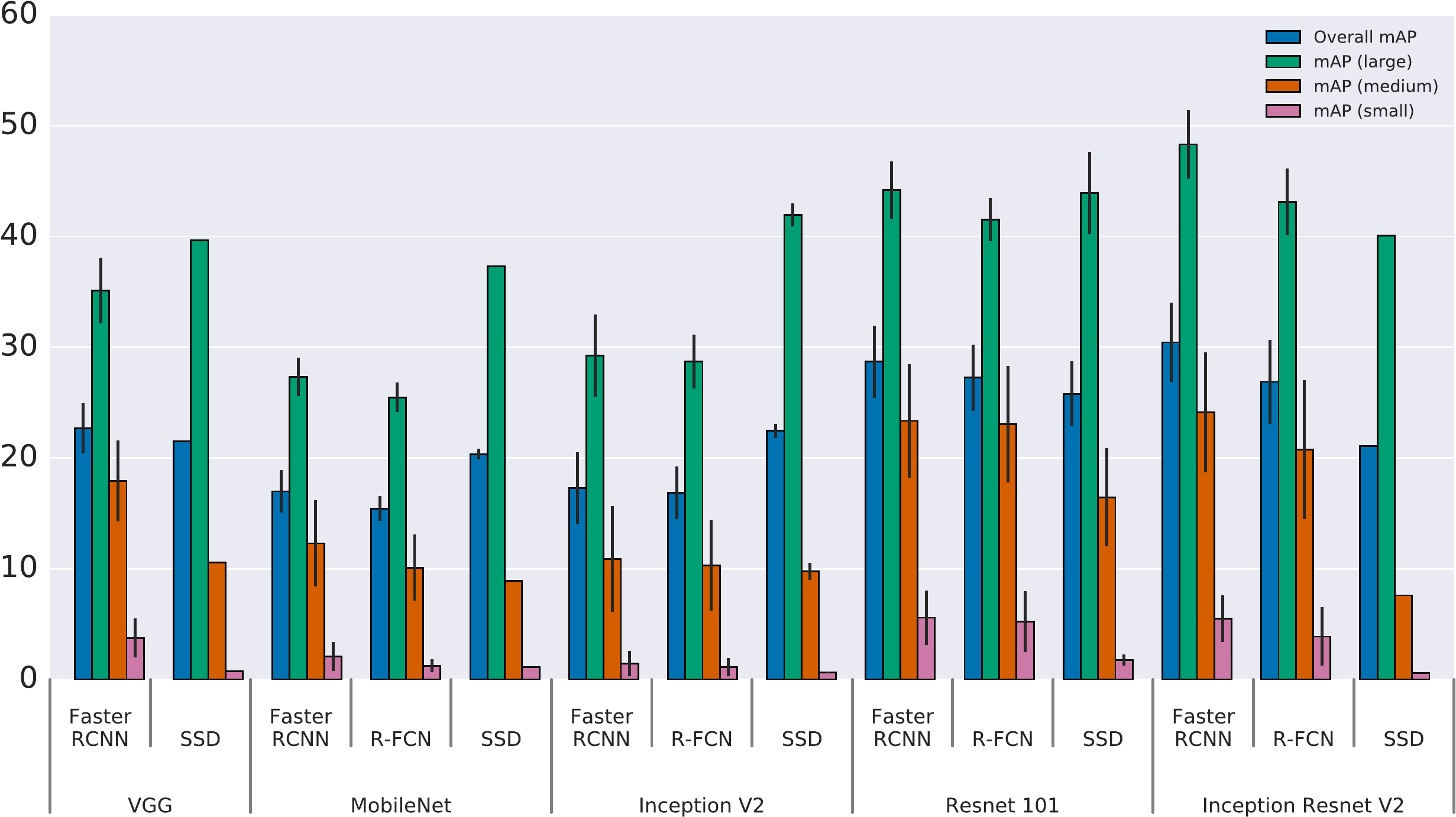}
\caption{
\footnotesize mAP for each object size, stratified by meta-architecture and feature extractor,
for image resolution of 300.
This is a larger version of Figure 4a from main paper.
}
\label{fig:map_by_object_size}
\end{center}\vspace{8mm}
\end{figure*}

\begin{figure*}[t!]
\begin{center}
\includegraphics[
	width=.9\linewidth]{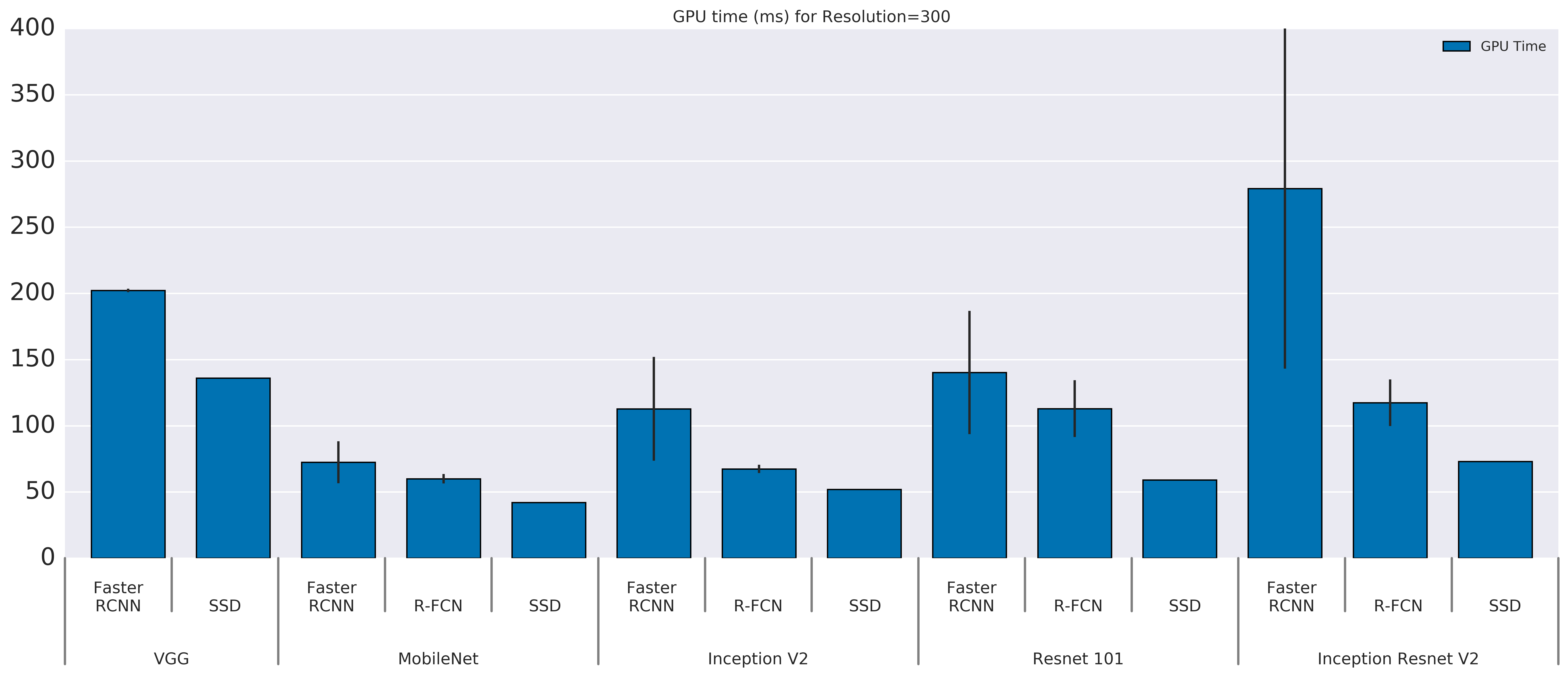}
\caption{
\footnotesize GPU time (milliseconds), stratified by meta-architecture and feature extractor,
for image resolution of 300.
}
\label{fig:bar_gpu_time}
\end{center}\vspace{8mm}
\end{figure*}

\paragraph{``Bang-for-your-buck''.}
Figure~\ref{fig:map_by_object_size} plots the mAP for each architecture and feature extractor,
and Figure~\ref{fig:bar_gpu_time} plots the corresponding GPU time (for images of size 300).
We also measure ``bang-for-your-buck'' in Figures~\ref{fig:map_per_gputime} (using GPU wallclock time)
and~\ref{fig:map_per_cputime} (using CPU wallclock time), plotting the 
mAP points per millisecond of computation time for each detector.
The general trend that we observe is that SSD has the highest mAP per millisecond, followed by R-FCN, then by Faster R-CNN models.  The bang-for-your-buck for SSD with Mobilenet is particularly notable on large objects, achieving about 0.9 mAP points per millisecond of computation on a GPU.

\begin{figure*}[t!]
\begin{center}
\includegraphics[
	width=.9\linewidth]{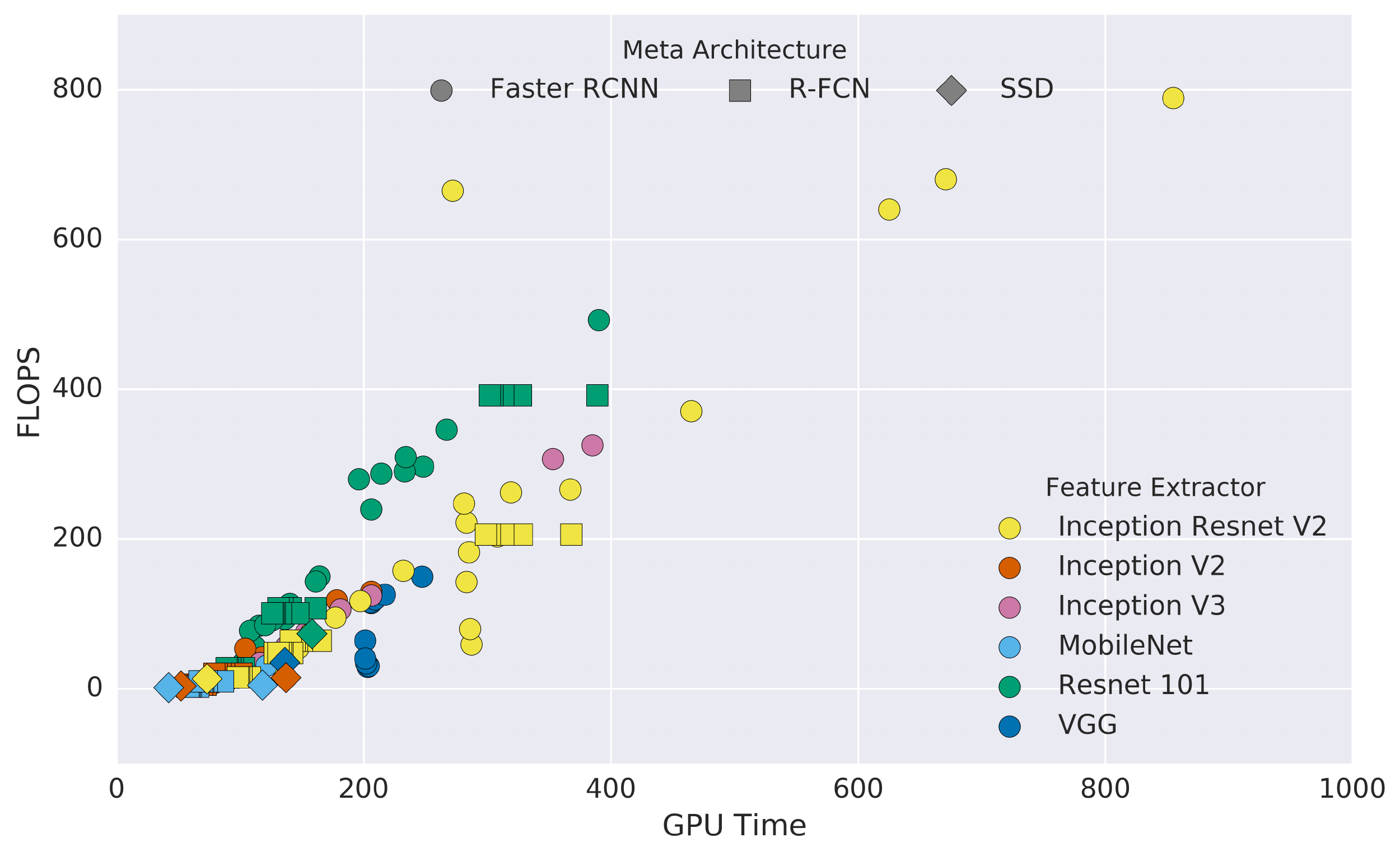}
\caption{
\footnotesize FLOPS vs. GPU wallclock time.
}
\label{fig:flops_v_gputime}
\end{center}\vspace{8mm}
\end{figure*}

\begin{figure*}[t!]
\begin{center}
\includegraphics[
	width=.9\linewidth]{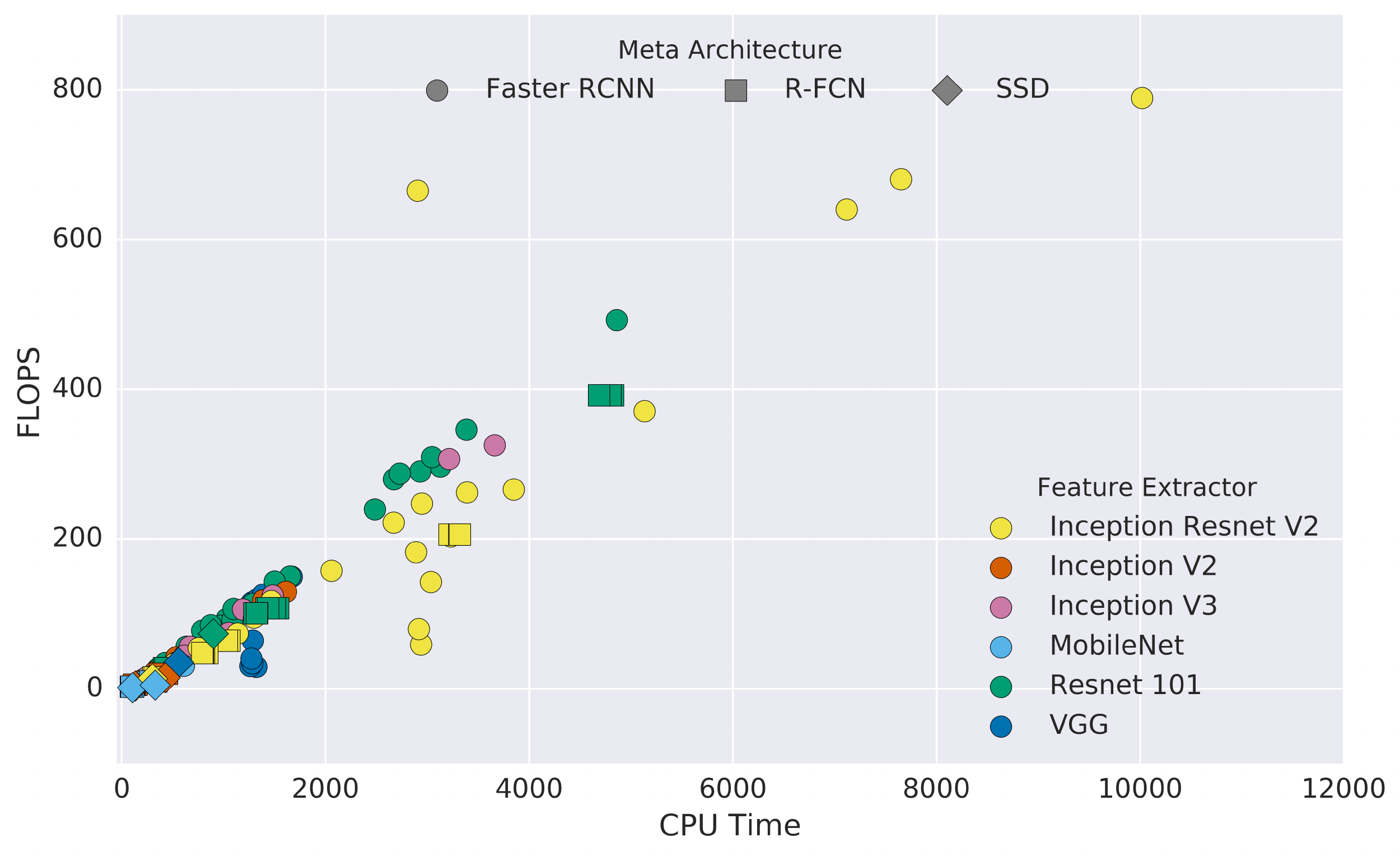}
\caption{
\footnotesize FLOPS vs. CPU wallclock time.
}
\label{fig:flops_v_cputime}
\end{center}
\end{figure*}

\paragraph{FLOPs analysis.}
Counting FLOPs (multiply-adds)
gives us a platform independent measure of computation,
which may or may not be linear with respect to actual running times
due to a number of issues such as caching, I/O, hardware optimization etc,  

Figures~\ref{fig:flops_v_gputime} and~\ref{fig:flops_v_cputime} plot
the FLOP count against observed wallclock times on the GPU and CPU respectively.
Interestingly,  we observe in the GPU plot (Figure~\ref{fig:flops_v_gputime})
that each model has a different average ratio of flops to observed running time in milliseconds.  For denser block models such as Resnet 101, FLOPs/GPU time is typically greater than 1, perhaps due to efficiency in caching. For Inception and Mobilenet models, this ratio is typically less than 1 --- we conjecture that this could be that factorization reduces FLOPs, but adds more overhead in memory I/O or potentially that current GPU instructions (cuDNN) are more optimized for dense convolution.

\begin{figure*}[t!]
\begin{center}
\includegraphics[
	width=.75\linewidth]{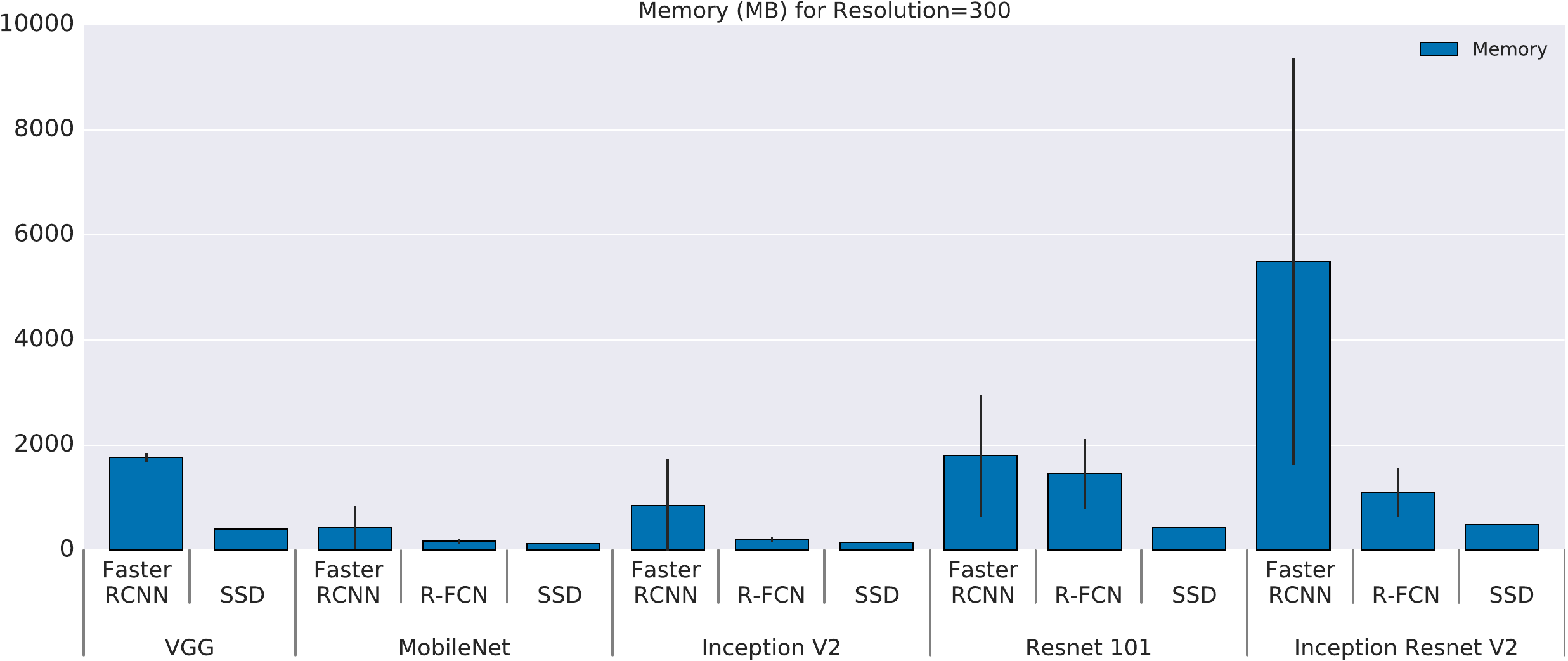}
\caption{
\footnotesize Memory (Mb) barplot.  Note that we measure total memory usage
rather than peak memory usage.  Moreover, we include all data points corresponding to the low-resolution models here.  The error bars reflect
variance in memory usage by using different numbers of proposals for the Faster R-CNN and R-FCN models (which leads to the seemingly
considerable variance in the Faster-RCNN with Inception Resnet bar).
}
\label{fig:memory_barplot}
\end{center}
\end{figure*}

\paragraph{Memory analysis.}
For memory benchmarking, we measure total usage rather than peak  usage.  Figures~\ref{fig:flops_v_gputime},~\ref{fig:flops_v_cputime} plot memory usage against GPU and CPU wallclock times.  Overall, we observe high correlation with running time with larger and more powerful feature extractors requiring much more memory. Figure~\ref{fig:memory_barplot} plots some of the same information in more detail, drilling down by meta-architecture and feature extractor selection.  As with speed, Mobilenet is again the cheapest, requiring less than 1Gb (total) memory in almost all settings.
\paragraph{Good localization at .75 IOU means good localization at all IOU thresholds.}

\begin{figure*}[t!]
\begin{center}
\includegraphics[
	width=.9\linewidth]{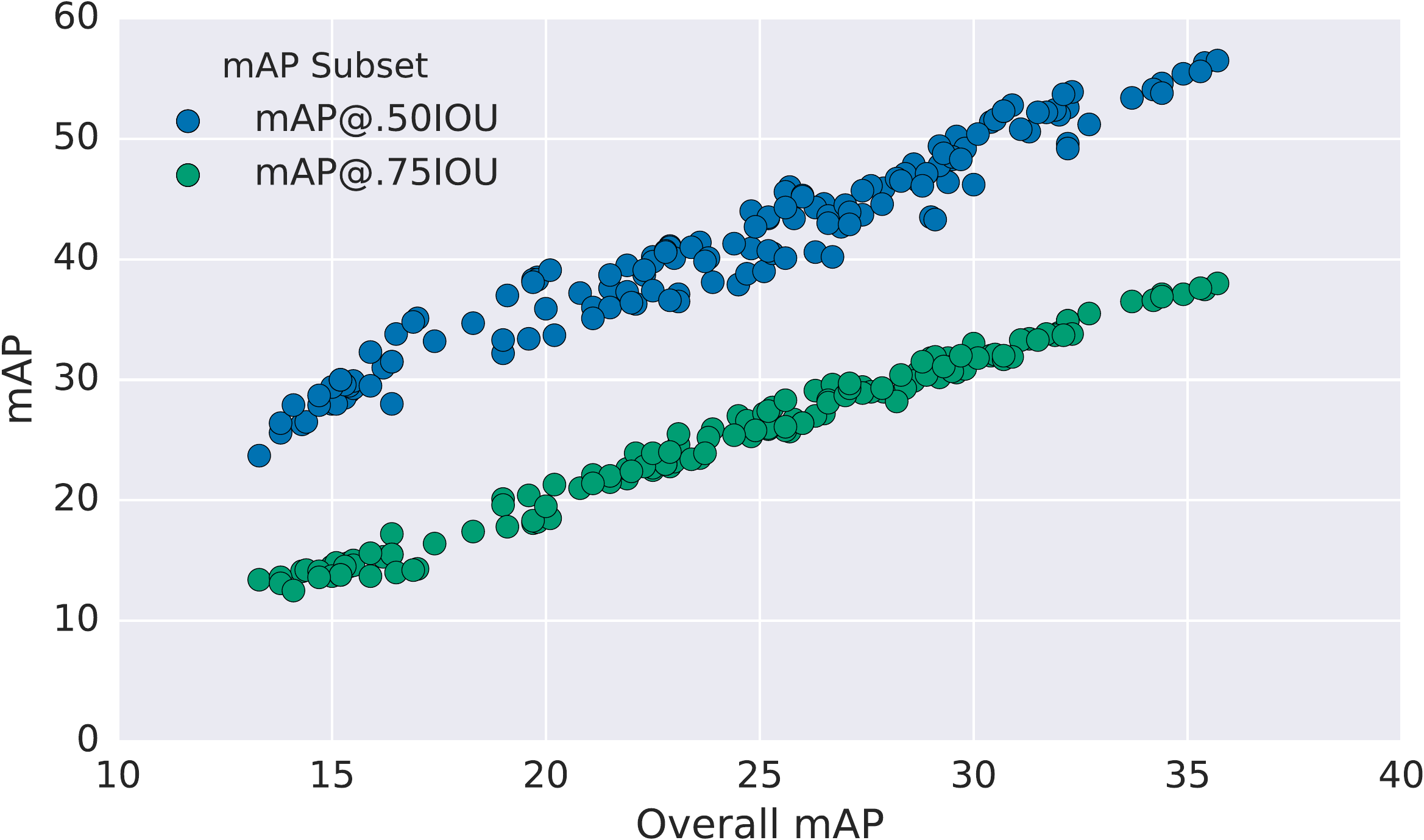}
\caption{
\footnotesize Overall COCO mAP (@[.5:.95]) for all experiments plotted against
corresponding mAP@.50IOU and mAP@.75IOU.  It is unsurprising that these numbers
are correlated, but it is interesting that they are almost perfectly correlated so for these models, it is never the case that a model has strong 
performance at 50\% IOU but weak performance at 75\% IOU.
}
\label{fig:map_50_v_75}
\end{center}
\end{figure*}

While slicing the data by object size leads to interesting insights, it is also worth nothing that slicing data by IOU threshold does not give much additional information.  Figure~\ref{fig:map_50_v_75} shows in fact that both  mAP@.5 and mAP@.75 performances are almost perfectly linearly correlated with mAP@[.5:.95].  Thus detectors that have poor performance at the higher IOU thresholds always also show poor performance at the lower IOU thresholds.
This being said, we also observe that mAP@.75 is slightly more tightly correlated with mAP@[.5:.95] (with $R^2>.99$), so if we were to replace the standard COCO metric with mAP at a single IOU threshold, we would likely choose IOU=.75.
\paragraph{Detectors on the Optimality Frontier: Visualizations and test-dev numbers.}
Finally in the remaining figures, we visualize detections on images from the COCO dataset, showing side-by-side comparisons of five
of the detectors that lie on the ``optimality frontier'' of the speed-accuracy
trade-off plot.  To visualize, we select detections with score greater than a  threshold and plot the top 20 detections in each image.  We use a threshold of .5 for Faster R-CNN and R-FCN and .3 for SSD.  These thresholds were hand-tuned for (subjective) visual attractiveness and not using  rigorous criteria so we caution viewers from reading too much into the tea leaves from these visualizations.  This being said, we see that across our examples, all of the detectors perform reasonably well on large objects --- SSD only shows its weakness on small objects, missing some of the smaller kites and people in the first image as well as the smaller cups and bottles on the dining table in the last image.

Due to having too many models to evaluate, we report results on a held out 
``minival'' set, checking against the test-dev set 
only on a subset of models.  Table~\ref{tab:testdev} compares mAP on 
the minival set against that on test-dev on the same optimality frontier points that we have visualized, and we see that there is good agreement for both datasets.

\begin{figure*}[t!]
\begin{center}
\includegraphics[
	width=.9\linewidth]{Figures/map_per_gputime.pdf}
\caption{
\footnotesize ``Bang-for-your-buck'': mAP points per unit gpu wallclock time (ms). 
}
\label{fig:map_per_gputime}
\end{center}\vspace{8mm}
\end{figure*}

\begin{figure*}[t!]
\begin{center}
\includegraphics[
	width=.9\linewidth]{Figures/map_per_cputime.pdf}
\caption{
\footnotesize ``Bang-for-your-buck'': mAP points per unit cpu wallclock time (ms). 
}
\label{fig:map_per_cputime}
\end{center}
\end{figure*}

\begin{figure*}[t!]
\begin{center}
\includegraphics[
	width=.9\linewidth]{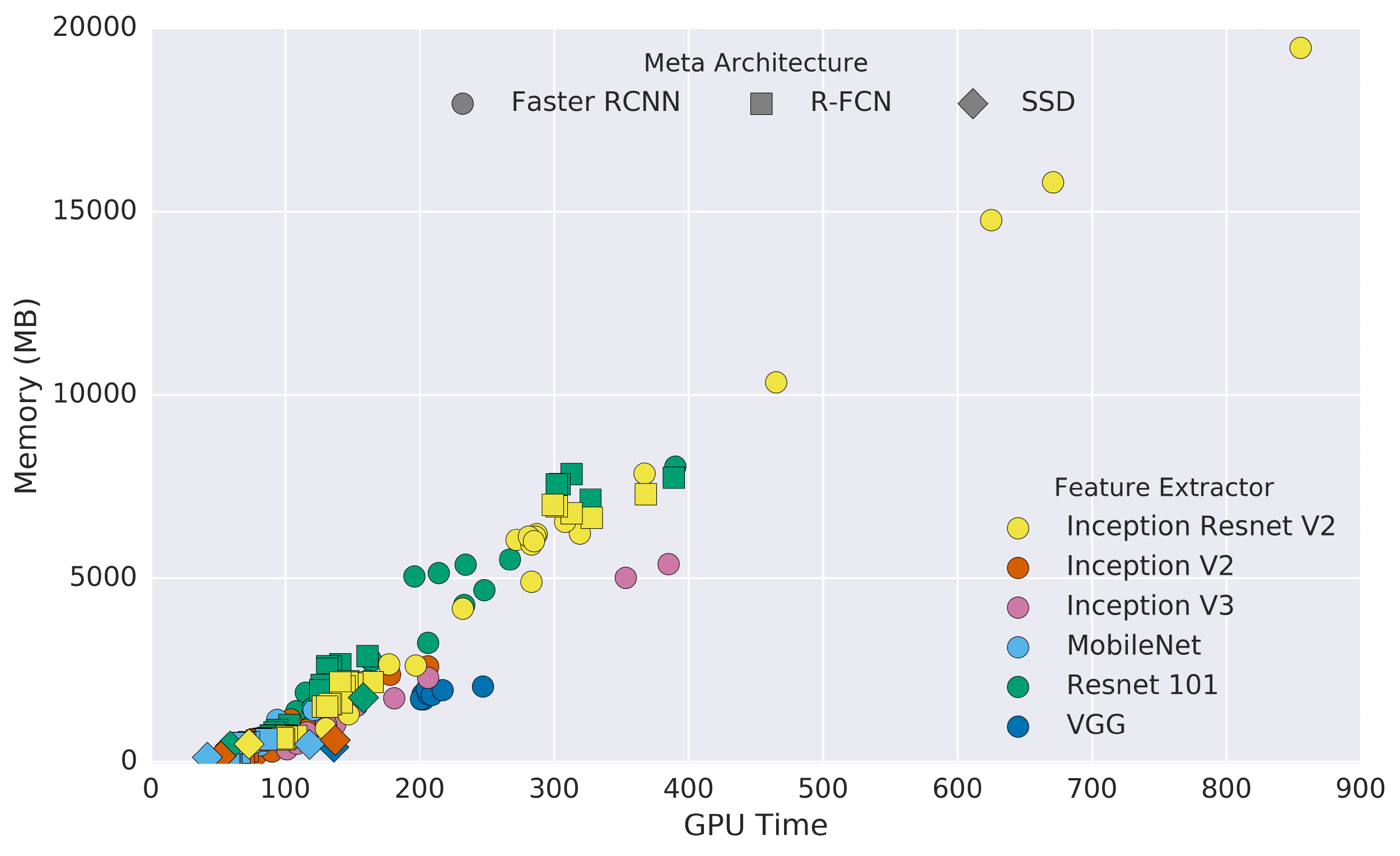}
\caption{
\footnotesize Memory (Mb) vs. GPU wallclock time.
}
\label{fig:memory_v_gputime}
\end{center}
\end{figure*}

\begin{figure*}[t!]
\begin{center}
\includegraphics[
	width=.9\linewidth]{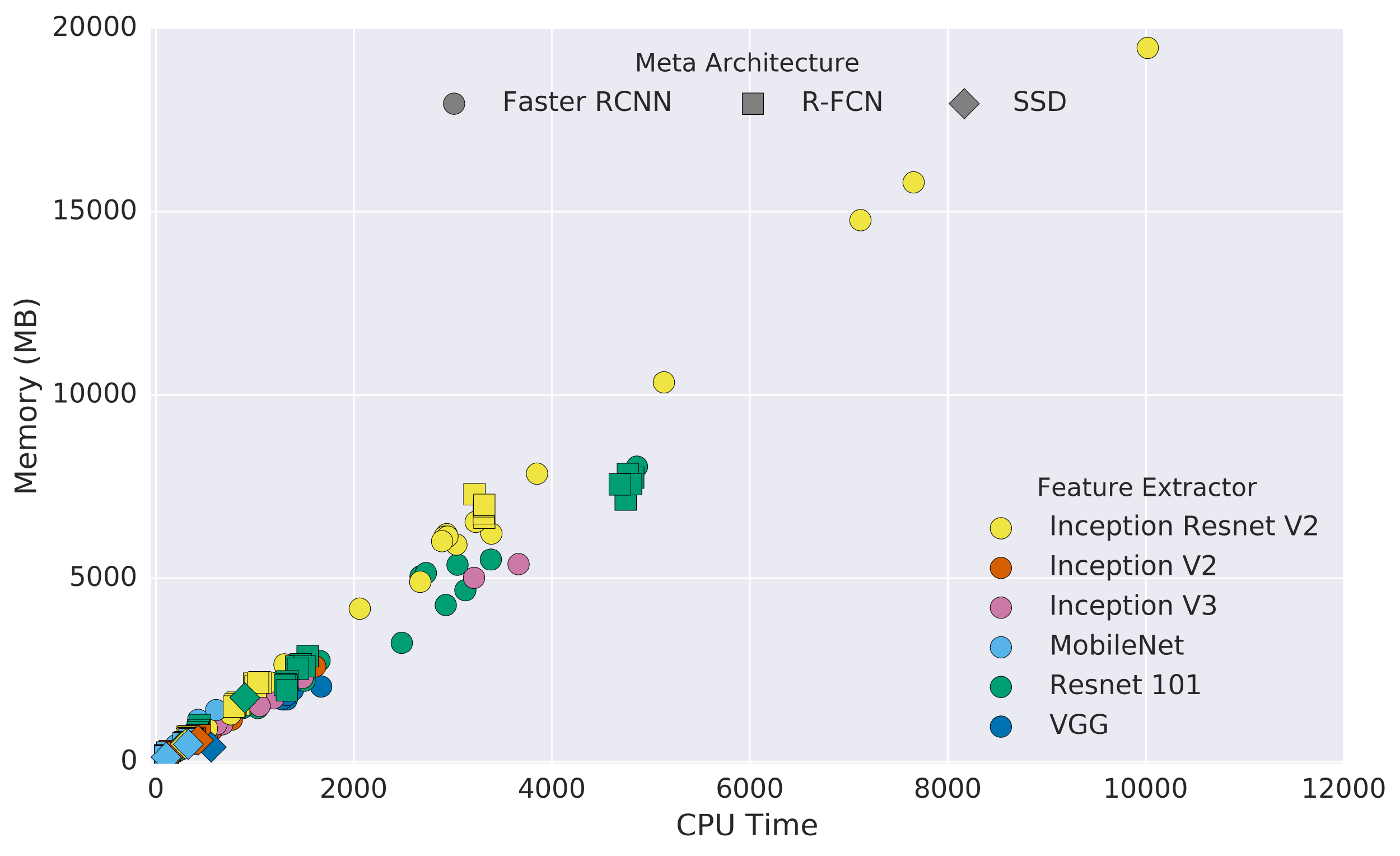}
\caption{
\footnotesize Memory (Mb) vs. CPU wallclock time.
}
\label{fig:memory_v_cputime}
\end{center}
\end{figure*}

\begin{figure*}[t!]
\captionsetup[subfigure]{justification=centering}
\begin{center}
\subfloat[SSD w/MobileNet]{
\setlength{\fboxsep}{0pt}%
\setlength{\fboxrule}{1pt}%
\fbox{\includegraphics[height=32mm]{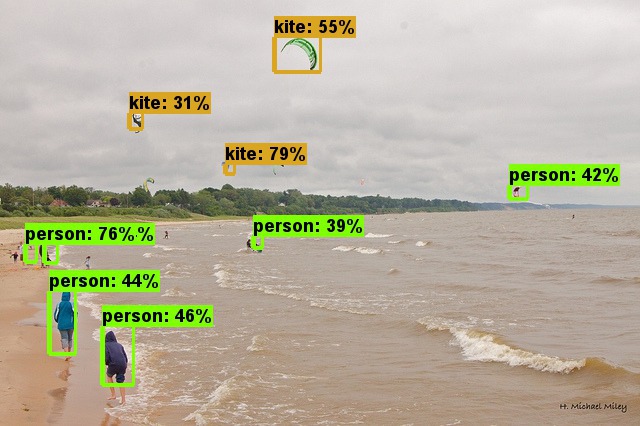}}
}\qquad
\subfloat[SSD w/Inception V2]{
\setlength{\fboxsep}{0pt}%
\setlength{\fboxrule}{1pt}%
\fbox{\includegraphics[height=32mm]{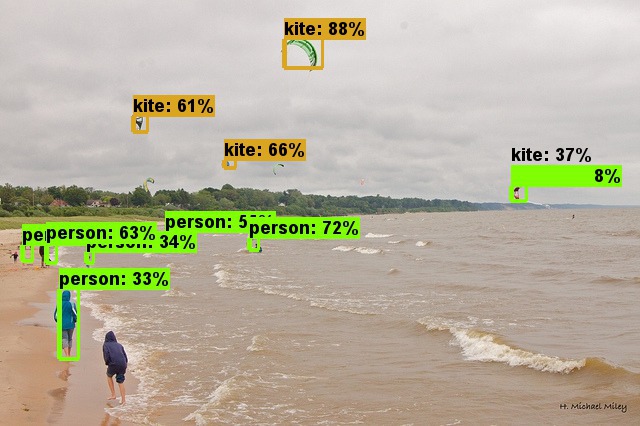}}
}\qquad
\subfloat[Faster R-CNN w/Resnet 101, 100 Proposals]{
\setlength{\fboxsep}{0pt}%
\setlength{\fboxrule}{1pt}%
\fbox{\includegraphics[height=32mm]{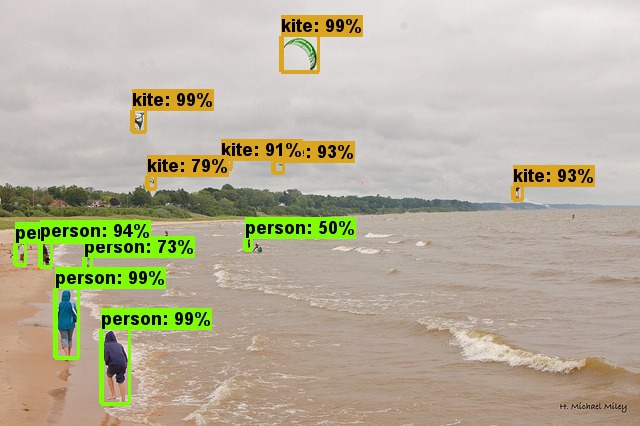}}
}\\
\subfloat[R-FCN w/Resnet 101, 300 Proposals]{
\setlength{\fboxsep}{0pt}%
\setlength{\fboxrule}{1pt}%
\fbox{\includegraphics[height=32mm]{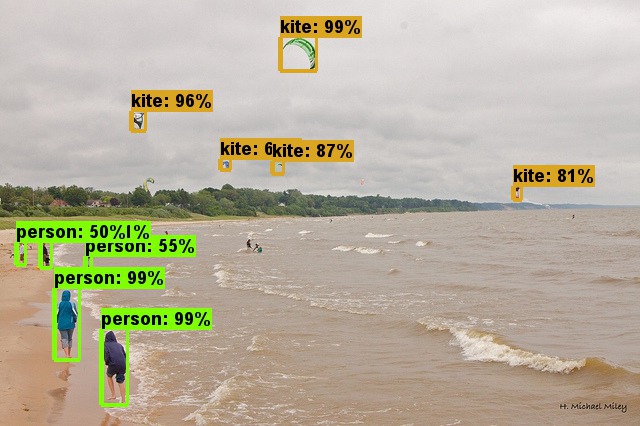}}
}\qquad
\subfloat[Faster R-CNN w/Inception Resnet V2, 300 Proposals]{
\setlength{\fboxsep}{0pt}%
\setlength{\fboxrule}{1pt}%
\fbox{\includegraphics[height=32mm]{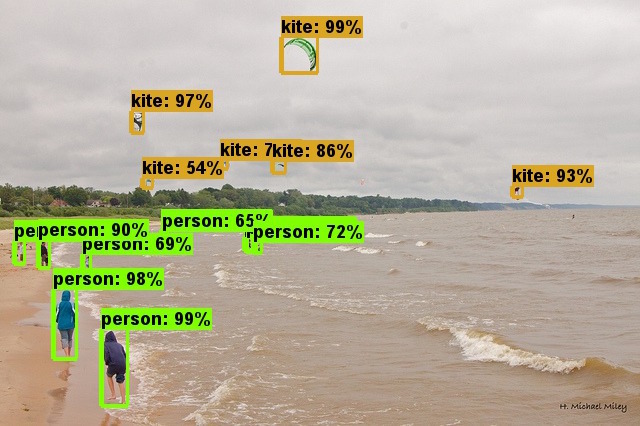}}
}
\end{center}
\end{figure*}

\begin{figure*}[t!]
\captionsetup[subfigure]{justification=centering}
\begin{center}
\subfloat[SSD w/MobileNet]{
\setlength{\fboxsep}{0pt}%
\setlength{\fboxrule}{1pt}%
\fbox{\includegraphics[height=50mm]{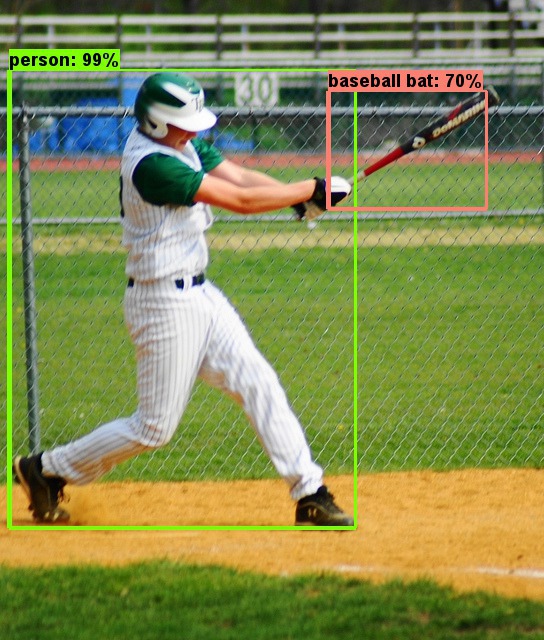}}
}\qquad
\subfloat[SSD w/Inception V2]{
\setlength{\fboxsep}{0pt}%
\setlength{\fboxrule}{1pt}%
\fbox{\includegraphics[height=50mm]{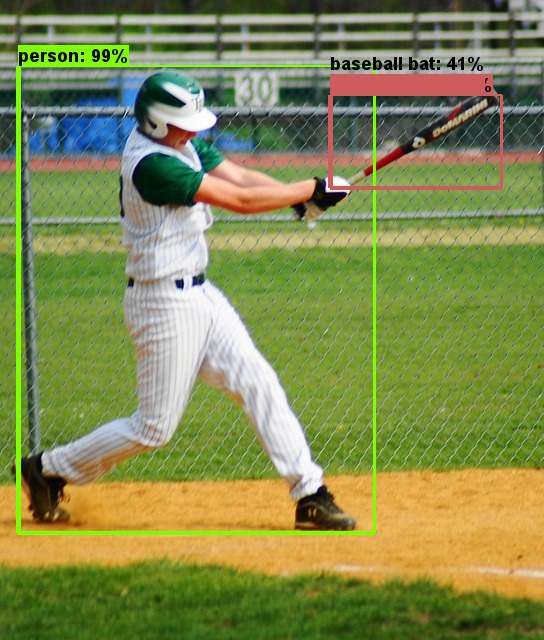}}
}\qquad
\subfloat[Faster R-CNN w/Resnet 101, 100 Proposals]{
\setlength{\fboxsep}{0pt}%
\setlength{\fboxrule}{1pt}%
\fbox{\includegraphics[height=50mm]{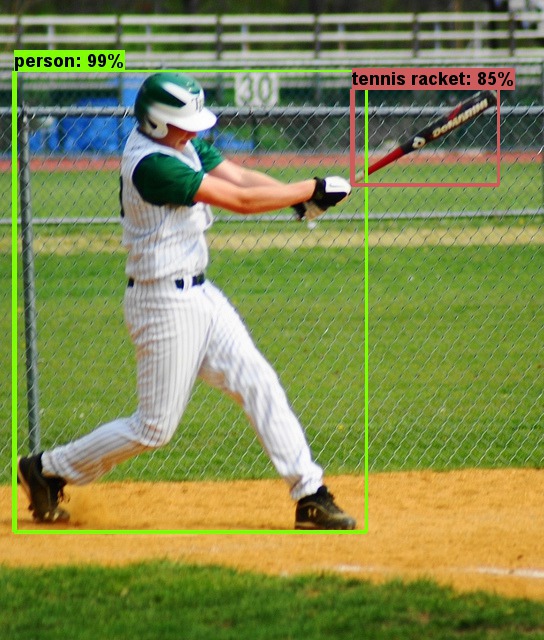}}
}\\
\subfloat[R-FCN w/Resnet 101, 300 Proposals]{
\setlength{\fboxsep}{0pt}%
\setlength{\fboxrule}{1pt}%
\fbox{\includegraphics[height=50mm]{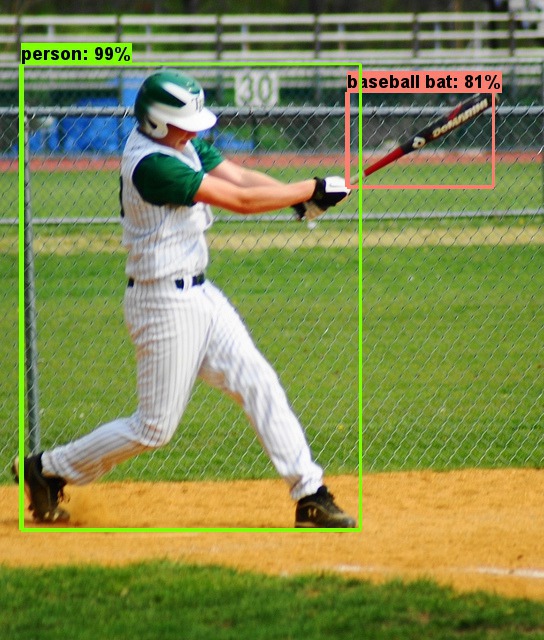}}
}\qquad
\subfloat[Faster R-CNN w/Inception Resnet V2, 300 Proposals]{
\setlength{\fboxsep}{0pt}%
\setlength{\fboxrule}{1pt}%
\fbox{\includegraphics[height=50mm]{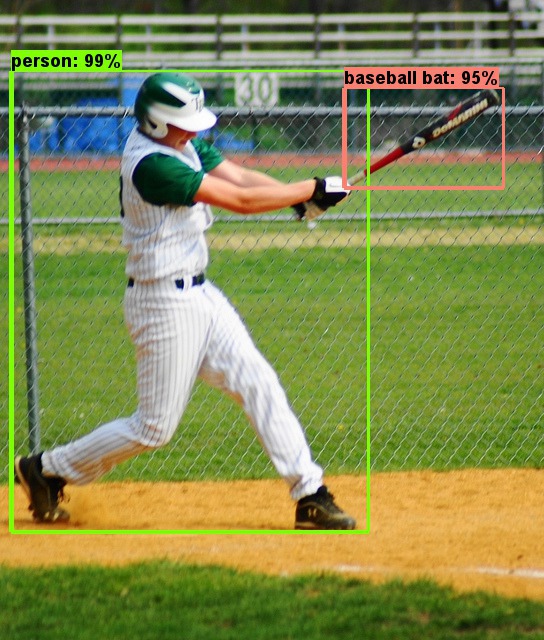}}
}
\end{center}
\end{figure*}

\begin{figure*}[t!]
\captionsetup[subfigure]{justification=centering}
\begin{center}
\subfloat[SSD w/MobileNet]{
\setlength{\fboxsep}{0pt}%
\setlength{\fboxrule}{1pt}%
\fbox{\includegraphics[height=32mm]{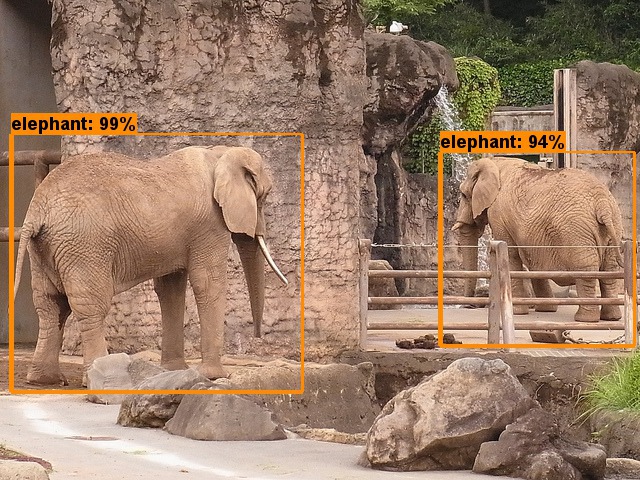}}
}\qquad
\subfloat[SSD w/Inception V2]{
\setlength{\fboxsep}{0pt}%
\setlength{\fboxrule}{1pt}%
\fbox{\includegraphics[height=32mm]{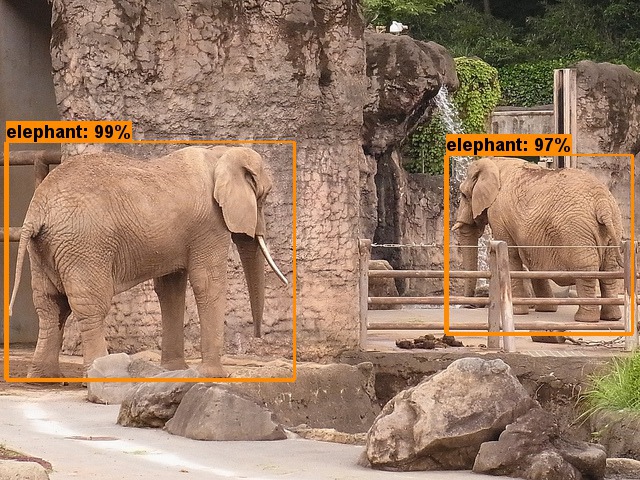}}
}\qquad
\subfloat[Faster R-CNN w/Resnet 101, 100 Proposals]{
\setlength{\fboxsep}{0pt}%
\setlength{\fboxrule}{1pt}%
\fbox{\includegraphics[height=32mm]{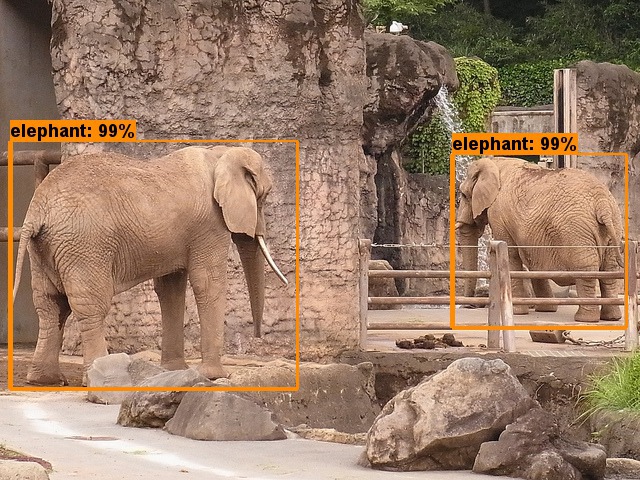}}
}\\
\subfloat[R-FCN w/Resnet 101, 300 Proposals]{
\setlength{\fboxsep}{0pt}%
\setlength{\fboxrule}{1pt}%
\fbox{\includegraphics[height=32mm]{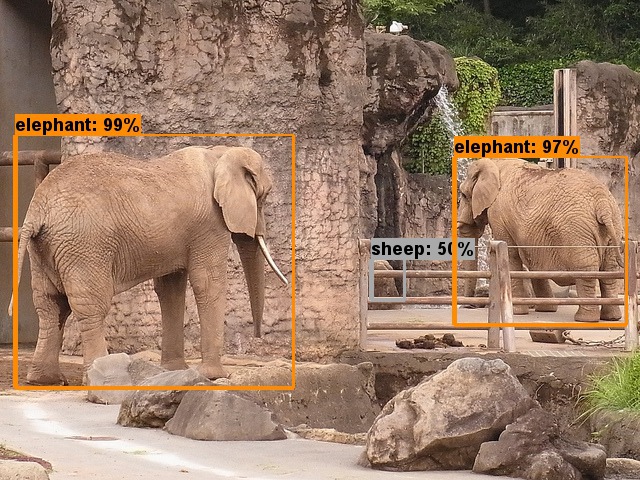}}
}\qquad
\subfloat[Faster R-CNN w/Inception Resnet V2, 300 Proposals]{
\setlength{\fboxsep}{0pt}%
\setlength{\fboxrule}{1pt}%
\fbox{\includegraphics[height=32mm]{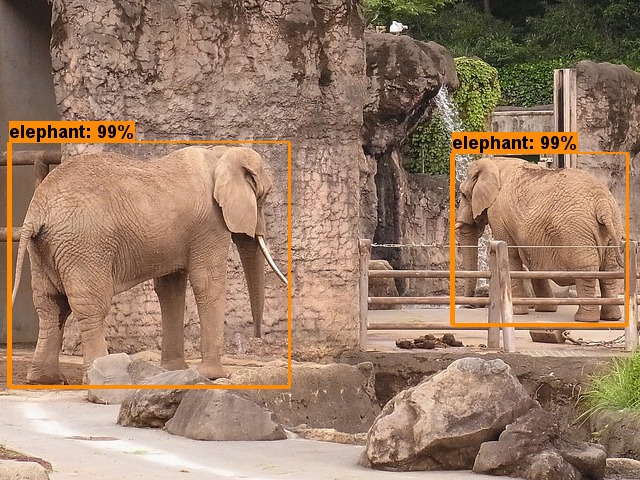}}
}
\end{center}
\end{figure*}

\begin{figure*}[t!]
\captionsetup[subfigure]{justification=centering}
\begin{center}
\subfloat[SSD w/MobileNet]{
\setlength{\fboxsep}{0pt}%
\setlength{\fboxrule}{1pt}%
\fbox{\includegraphics[height=50mm]{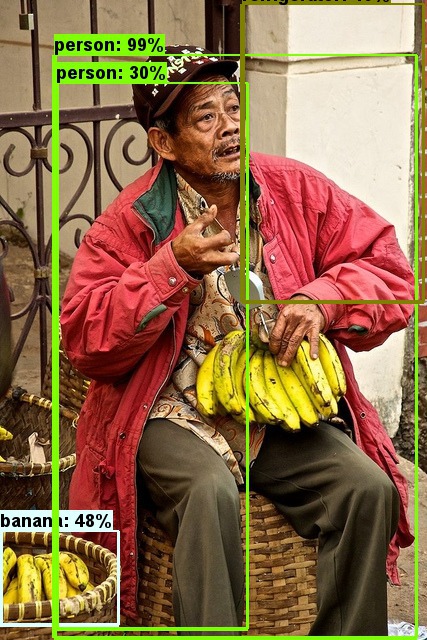}}
}\qquad
\subfloat[SSD w/Inception V2]{
\setlength{\fboxsep}{0pt}%
\setlength{\fboxrule}{1pt}%
\fbox{\includegraphics[height=50mm]{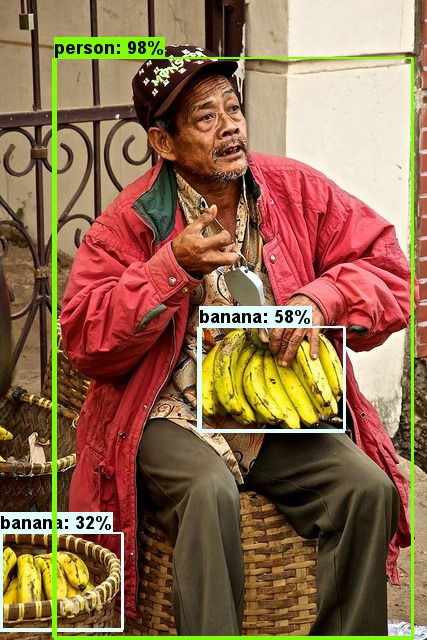}}
}\qquad
\subfloat[Faster R-CNN w/Resnet 101, 100 Proposals]{
\setlength{\fboxsep}{0pt}%
\setlength{\fboxrule}{1pt}%
\fbox{\includegraphics[height=50mm]{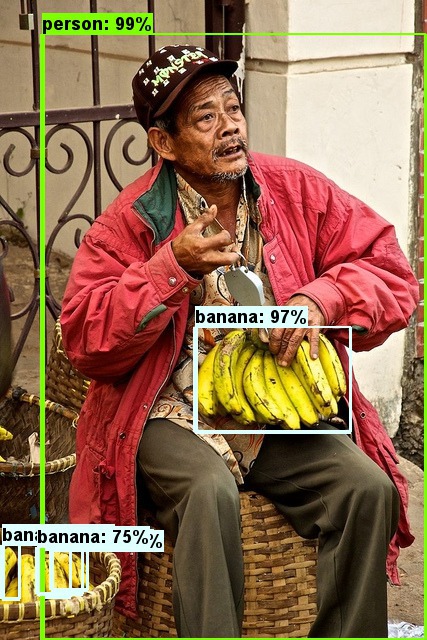}}
}\\
\subfloat[R-FCN w/Resnet 101, 300 Proposals]{
\setlength{\fboxsep}{0pt}%
\setlength{\fboxrule}{1pt}%
\fbox{\includegraphics[height=50mm]{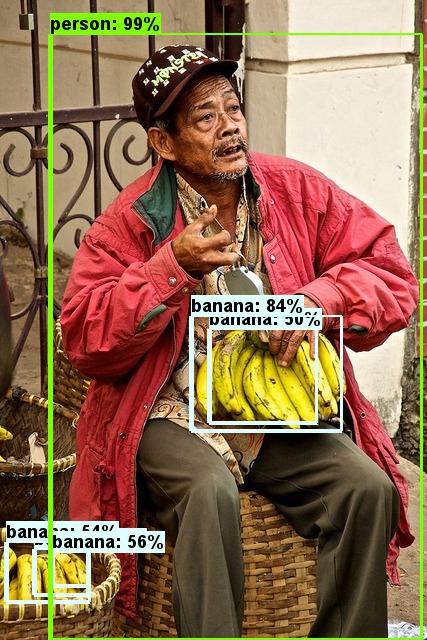}}
}\qquad
\subfloat[Faster R-CNN w/Inception Resnet V2, 300 Proposals]{
\setlength{\fboxsep}{0pt}%
\setlength{\fboxrule}{1pt}%
\fbox{\includegraphics[height=50mm]{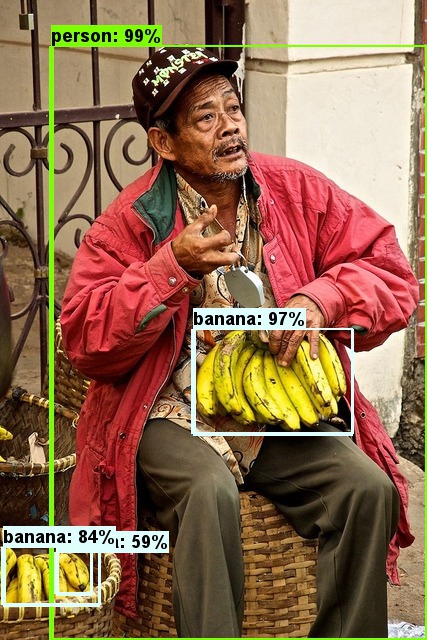}}
}
\end{center}
\end{figure*}

\begin{figure*}[t!]
\captionsetup[subfigure]{justification=centering}
\begin{center}
\subfloat[SSD w/MobileNet]{
\setlength{\fboxsep}{0pt}%
\setlength{\fboxrule}{1pt}%
\fbox{\includegraphics[height=32mm]{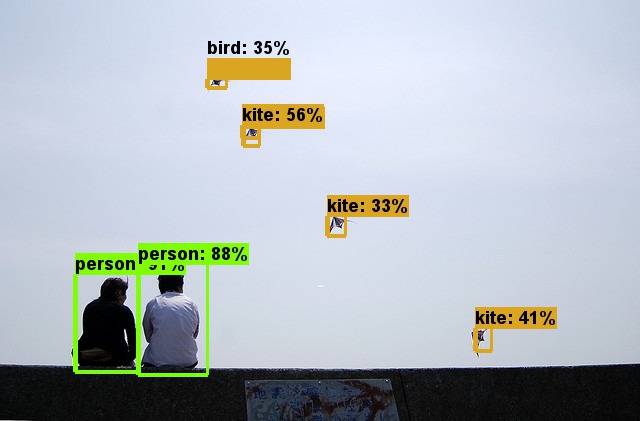}}
}\qquad
\subfloat[SSD w/Inception V2]{
\setlength{\fboxsep}{0pt}%
\setlength{\fboxrule}{1pt}%
\fbox{\includegraphics[height=32mm]{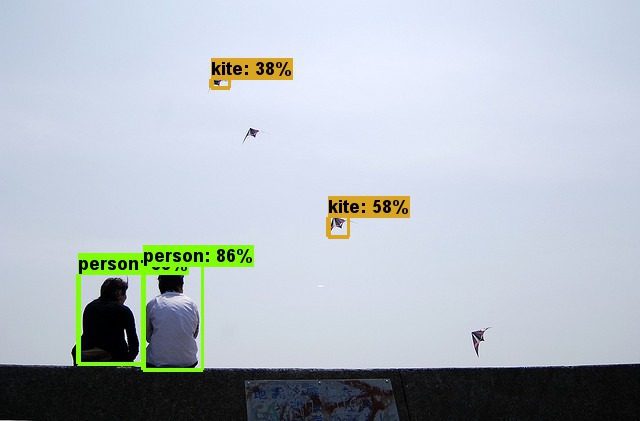}}
}\qquad
\subfloat[Faster R-CNN w/Resnet 101, 100 Proposals]{
\setlength{\fboxsep}{0pt}%
\setlength{\fboxrule}{1pt}%
\fbox{\includegraphics[height=32mm]{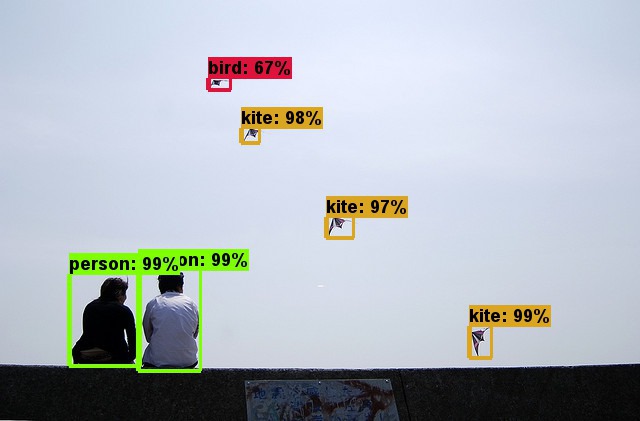}}
}\\
\subfloat[R-FCN w/Resnet 101, 300 Proposals]{
\setlength{\fboxsep}{0pt}%
\setlength{\fboxrule}{1pt}%
\fbox{\includegraphics[height=32mm]{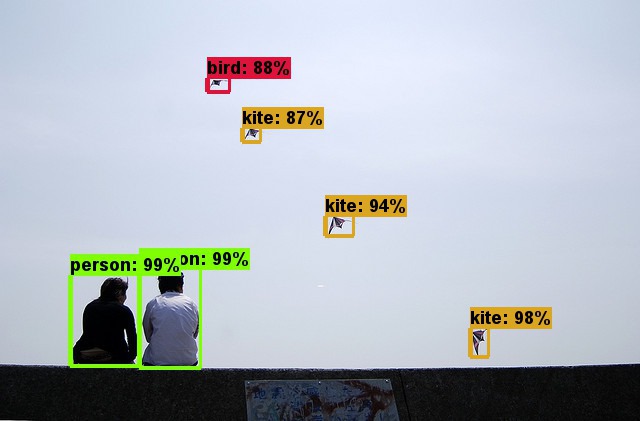}}
}\qquad
\subfloat[Faster R-CNN w/Inception Resnet V2, 300 Proposals]{
\setlength{\fboxsep}{0pt}%
\setlength{\fboxrule}{1pt}%
\fbox{\includegraphics[height=32mm]{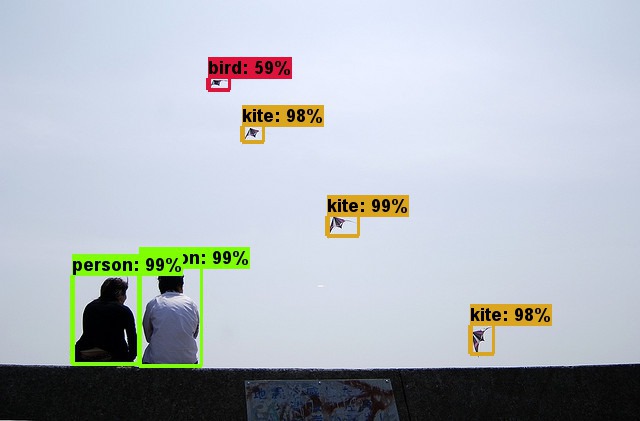}}
}
\end{center}
\end{figure*}

\begin{figure*}[t!]
\captionsetup[subfigure]{justification=centering}
\begin{center}
\subfloat[SSD w/MobileNet]{
\setlength{\fboxsep}{0pt}%
\setlength{\fboxrule}{1pt}%
\fbox{\includegraphics[height=32mm]{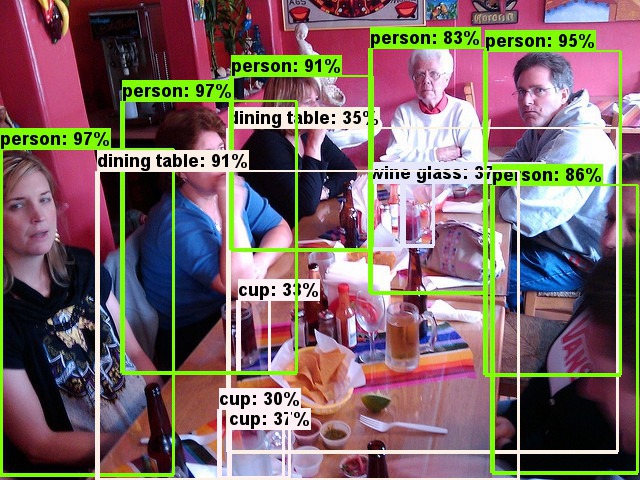}}
}\qquad
\subfloat[SSD w/Inception V2]{
\setlength{\fboxsep}{0pt}%
\setlength{\fboxrule}{1pt}%
\fbox{\includegraphics[height=32mm]{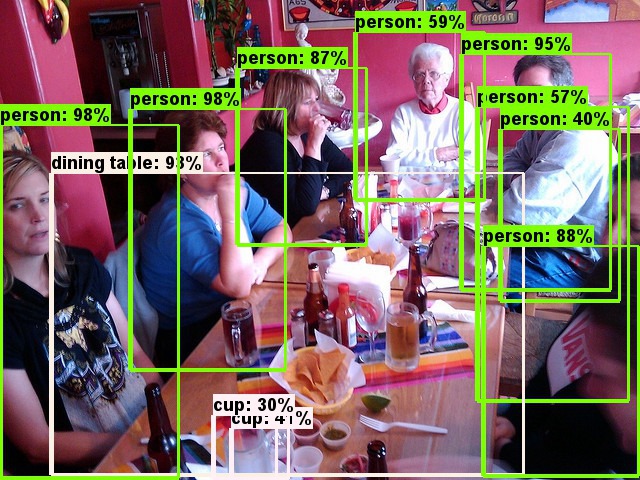}}
}\qquad
\subfloat[Faster R-CNN w/Resnet 101, 100 Proposals]{
\setlength{\fboxsep}{0pt}%
\setlength{\fboxrule}{1pt}%
\fbox{\includegraphics[height=32mm]{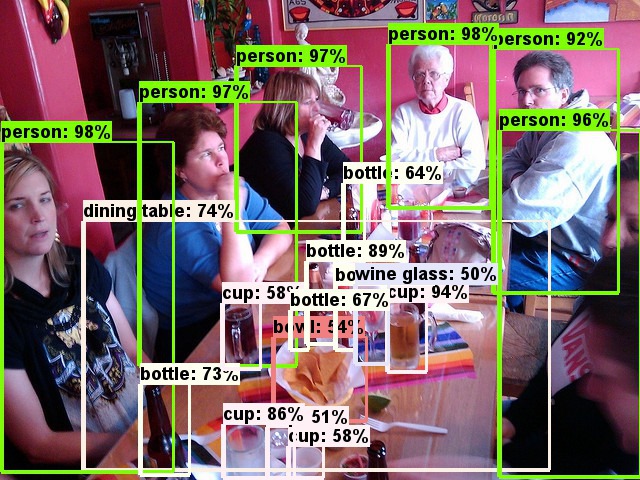}}
}\\
\subfloat[R-FCN w/Resnet 101, 300 Proposals]{
\setlength{\fboxsep}{0pt}%
\setlength{\fboxrule}{1pt}%
\fbox{\includegraphics[height=32mm]{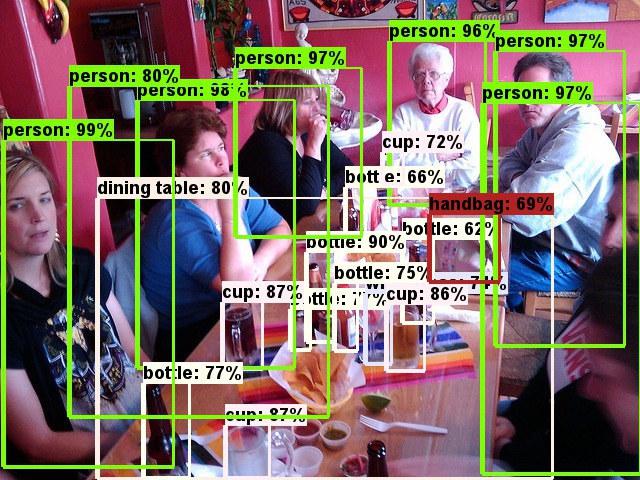}}
}\qquad
\subfloat[Faster R-CNN w/Inception Resnet V2, 300 Proposals]{
\setlength{\fboxsep}{0pt}%
\setlength{\fboxrule}{1pt}%
\fbox{\includegraphics[height=32mm]{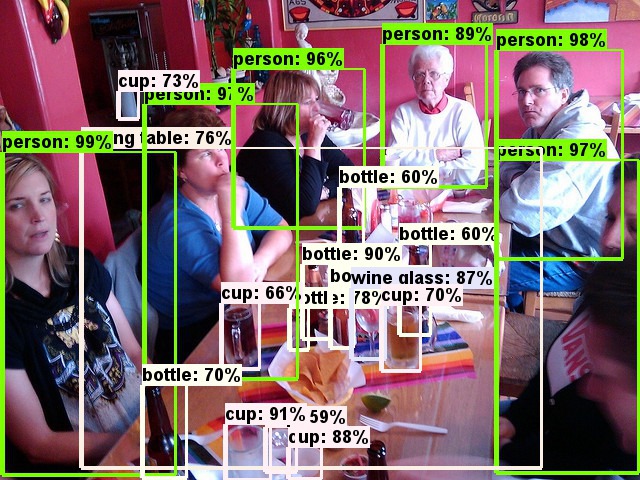}}
}
\end{center}
\end{figure*}